\documentclass[letterpaper]{article}
\usepackage[preprint]{aaai2027}
\usepackage[hyphens]{url}
\usepackage{graphicx}
\usepackage{natbib}
\usepackage{caption}
\usepackage{amsmath}
\usepackage{amssymb}
\usepackage{booktabs}
\usepackage{multirow}

\title{ALIVE: Warnings Before Exclusion in Budgeted Multi-Source Learning}

\author{
    Xiyang Zhang\textsuperscript{\rm 1,\rm 2},
    Hongzhi Wang\textsuperscript{\rm 1}\corresponding,
    Yuanhe Tian\textsuperscript{\rm 2}
}

\affiliations{
    \textsuperscript{\rm 1}Harbin Institute of Technology\\
    \textsuperscript{\rm 2}Zhongguancun Academy\\
    s-zhangxiy24@bza.edu.cn,
    wangzh@hit.edu.cn,
    tianyuanhe@zgci.ac.cn
}

\begin{document}

\maketitle

\begin{abstract}
A routing decision can be revised at the next transaction, but a latched source exclusion persists across later decisions.  We ask what evidence should authorize these unequal-persistence actions when finite-population auditing and learning share a budget.  ALIVE (\emph{Action-Layered Intervention via Evidence}) is an auditable control layer: one randomized without-replacement prefix supplies cached evidence, heuristic warnings drive non-latching
floor-bounded routing, and only two fresh simultaneous certificate separations may latch an exclusion request subject to capacity-feasible activation. Conditional on fixed support and labels under an ideal uniform audit permutation, any predictable controller preserving this interface inherits an anytime familywise bound of \(\delta\) on acting against a source that fails the pre-fixed absolute or relative strict-majority-disagreement predicate.  With a published known-size, all-strict-majority PPR engine, median evidence count fell from 304 to 96 identities in e40 and from 171 to 62 in e60, while both engines used 48 in e80.  In the matched CIFAR controller, the persistent-action layer added \(+0.1935\) accuracy-AUBC percentage points over routing-only in all ten paired seed clusters.  The \(+0.1954\)-point full-system contrast against CBR was also positive but did not meet the predeclared multiplicity-adjusted criterion (conditional Holm-adjusted sign-flip reference value =.097656).  On a fixed natural panel, exploratory PPR used a median closure prefix of 95 rather than 105 for exploratory Serfling/FPC, but still exposed 88.0\% of the panel and had no downstream task.  Together these results map a restraint--power--cost--utility boundary: the action contract controls a defined persistent decision, while net value depends on evidence margin, audit cost, and budget regime.
\end{abstract}

\section{Introduction}
\label{sec:introduction}

In a crowd-labeled or repeated-label learner, reducing an annotator's share for
the next batch is reversible; excluding that annotator from future optimization
is not.  The latter changes subsequent data availability and source
opportunity, even when a capacity check occasionally suppresses the exclusion.
A noisy score should therefore not silently escalate into a persistent action.
We study the evidence-to-action interface needed to separate these decisions
when auditing and learning draw from one budget.

This distinction exposes three questions.  \emph{Statistical validity} asks
whether evidence supports a pre-specified action predicate.  \emph{Semantic
validity} asks whether that predicate identifies error or harm.  \emph{Decision
validity} asks whether acting yields net downstream value after audit cost.
ALIVE controls the first question for a finite-population majority-disagreement
target and evaluates the latter two separately; it does not treat disagreement
as ground truth.  Our setting is the shared-identity, multi-annotator case of
multi-source learning, not disjoint domain streams or federated clients.

ALIVE (\emph{Action-Layered Intervention via Evidence}) makes this separation
operational.  It consumes one randomized, without-replacement prefix of the
complete all-source identity intersection.  A heuristic warning may alter
non-latching, floor-bounded routing and request more audit.  A persistent
request requires simultaneous source-by-prefix confidence separation on two
fresh evidence advances, and activation additionally requires that remaining
sources can fill the optimizer batch.  Evidence acquired during a decision is
not action-defining until the next decision.

Conditional on fixed support and potential labels under an ideal uniform audit
permutation, any predictable controller that preserves this prefix and
certificate interface inherits an anytime familywise bound of \(\delta\) on
persistent action against a source that fails either the pre-fixed absolute-threshold
or relative strict-majority-disagreement condition.  Warning quality is outside
this statement; changing routing can change evidence delay but not the declared
false-action target.

\paragraph{Contributions.}
\textbf{Action-layered formulation:} we separate heuristic warnings, latched
requests, and capacity-feasible active exclusions as actions with different
persistence under one budget.  \textbf{Interface-preserving evidence contract:}
we specify the randomized-prefix, frozen-cache, latch, and timing invariants
under which predictable controllers inherit familywise false-persistent-action
control.  We also derive a schedule-conditional audit-to-latch bound and an
analytical direct-contrast strengthening of the evaluated coordinatewise
certificate.  \textbf{Cross-layer evaluation:} we measure null restraint,
evidence delay, panel exposure, matched persistent-action value, and end-to-end
operating regimes instead of treating certificate validity as a proxy for
learning benefit.

\section{Related Work}
\label{sec:related}

\paragraph{Source filtering and aggregation.}
Closest to a persistent source action are confidence-based worker eviction,
disagreement-based reputation against adversaries, and label-consistency
filtering before aggregation \cite{joglekar2013evaluating,
jagabathula2014reputation,li2024consistency}.  Those methods target worker
error, adversarial aggregation, or label purification.  ALIVE instead gives anytime familywise control of false persistent actions
for a declared finite-population disagreement
target and separates a non-latching warning from a persistent request.

Active learning with imperfect labelers jointly chooses instances and oracles,
estimates expertise, or trades quality against cost
\cite{donmez2008proactive,donmez2009efficiently,zheng2010noisy,
yan2011active,yan2012multiple,huang2017costeffective,
chakraborty2020asking,gao2020costaccuracy}.  Latent-label and repeated-label
models aggregate answers or estimate reliability
\cite{dawid1979observer,raykar2010crowds,sheng2008another,karger2014budget}.
ALIVE does not estimate gold labels or source accuracy; it uses stable shared
identities to bind a specified evidence event to actions of different duration.

\paragraph{Sequential finite-population evidence.}
Sampling-without-replacement concentration is classical
\cite{hoeffding1963probability,serfling1974probability}; dedicated confidence
sequences and gambling constructions can be tighter
\cite{waudbysmith2020confidence,ryu2024gambling}.  The base controller uses a
Hoeffding comparison with explicit alpha spending.  We derive a
strict-majority direct-contrast certificate that pathwise contains this base
certificate, and instantiate a prior--posterior ratio (PPR) alternative with a
uniform prior when the complete population size is known and every identity has
a strict majority.  These are
separate level-\(\delta\) engines.  On a common source--prefix path, the nested
direct/base certificate union equals the direct certificate; a union with
non-nested PPR requires a pre-fixed error split.  The concentration tools are established;
our contribution is the action interface, its inheritance result, and the
measurement of delay and downstream cost.

\paragraph{Risk-sensitive actions.}
Conservative exploration and selective prediction limit risky decisions
\cite{wu2016conservative,garcelon2020conservative,chow1970reject,
eliyaniv2010selective}.  Their constraints apply to rewards or predictions,
whereas ALIVE controls a finite-population source-action predicate and allows
capacity to suppress activation without clearing the evidence latch.

\paragraph{Budget-sensitive data selection.}
Coresets, gradient matching, uncertainty, and class balancing choose training
examples \cite{sener2018active,killamsetty2021gradmatch,settles2009active,
bengar2022classbalanced}; budget-dependent strategy selection motivates a
strong cheap incumbent \cite{hacohen2022budget,hacohen2023selectal,
zhang2023tailor}.  We therefore hold the selector fixed in the matched action
ablation.  Compute-aware selection charges selection and optimization
\cite{yin2025compute,wan2025computational}; our shared ledger has the same
accounting aim but is not a hardware or monetary cost model.  Robust losses act
after acquisition \cite{natarajan2013noisy,patrini2017robust,
han2018coteaching,li2020dividemix}; ALIVE changes source opportunity and
eligibility rather than correcting labels or estimating a noise transition.

\section{Setting and Shared Ledger}
\label{sec:setting}

We study shared-identity multi-annotator learning with \(S\geq3\) sources and zero-based decisions
\(t=0,1,\ldots\).  At decision \(t\), the learner chooses integer source counts
\(\mathbf n_t=(n_{1t},\ldots,n_{St})\), acquires candidate window \(W_t\),
selects training batch \(B_t\subseteq W_t\), and updates parameters.  An
observation contains features, a source label \(Y_s(i)\), an opaque source key
\(s\), and, when available, shared identity \(i\).  No action-defining branch
receives an environment name, corruption rate, clean-source flag, validation
outcome, or test metric.  Fixed labels, \(S\geq3\), and nonempty common support
are guarantee preconditions rather than facts inferred at runtime: the current
implementation does not verify label stability, and an empty intersection
disables certification instead of aborting the incumbent learner.

Every method obeys the declared accounting contract
\begin{equation}
 \sum_t\left(c_t^{\rm acquire}+c_t^{\rm score}+
 c_t^{\rm maintain}+c_t^{\rm train}\right)\leq\mathcal B .
 \label{eq:ledger}
\end{equation}
Complete audit groups occupy candidate slots and are charged once.  Entropy
ranking pays one forward pass, CBR no ranking pass, and Full-Consensus every
source query and tally.  Each method stops before exceeding \(\mathcal B\).
Thus ``same budget'' means one frozen reproducibility ledger, not equal labels,
gradient steps, seconds, FLOPs, energy, or money.

\section{Evidence-Gated Source Actions}
\label{sec:method}

\subsection{Randomized shared-identity prefix}

Let \(U\) be the fixed intersection of all sources' audit-eligible supports.
Before audit labels are read, the controller draws one domain-separated
pseudorandom permutation \(\pi\) of \(U\), idealized as uniform in the theory.
Each request consumes the next unused identities and admits only complete
all-source groups.  Adaptive request
sizes may change prefix length, but cannot select, revisit, or replace an
identity according to its labels.  Evidence acquired at decision \(t\) changes
the controller only at \(t+1\).

\paragraph{One-decision execution and audit.}
At decision \(t\), predicates are evaluated only from the prefix cached through
\(t-1\); the resulting states fix audit share, routing, integer allocation,
exclusion request, and eligibility.  The next complete groups and ordinary
candidates are then acquired.  New groups may enter the audit data structure,
but the frozen cache prevents them from changing action \(t\).  After scoring,
capacity fallback is recorded, the eligible batch is selected and trained, and
all transitions and charges are logged; new evidence can first act at \(t+1\).

A strict replay auditor checks four invariants before reading utility: admitted
identities are exactly one prefix with no duplicates or partial groups;
\textsc{Provisional} sources remain eligible; first latch requires two
certificate-positive fresh advances and can never clear; and recomputed event
charges reproduce an in-budget ledger.  These are trajectory facts, not
inferences from accuracy or a method name.

For identity \(i\), let \(m(i)\) be its unique strict cross-source majority
label; identities without a strict majority abstain.  If
\(a_1,\ldots,a_n\) are the first \(n\) comparable
identities induced by \(\pi\), define
\begin{equation}
 Z_s(i)=\mathbb I\{Y_s(i)\neq m(i)\},\qquad
 \widehat p_{s,n}=n^{-1}\sum_{k=1}^n Z_s(a_k).
 \label{eq:disagreement}
\end{equation}
The focal source participates in its majority.  This is not a leave-one-source-
out or truth estimator.

Before labels are read, the action designer fixes \(\tau\in(0,1)\).  ALIVE uses
\(\tau=1/S\) as a symmetric exact-null policy reference, not an estimated or
universal corruption rate or a consequence of majority voting.

\subsection{Warning, certificate, and actions}

After every source has at least eight comparable observations, source \(s\)
has the empirical warning
\begin{equation}
 W_{s,n}=\mathbb I\!\left\{\widehat p_{s,n}>\tau,\quad
 \widehat p_{s,n}>\frac{1}{S-1}\sum_{j\neq s}\widehat p_{j,n}\right\}.
 \label{eq:warning}
\end{equation}
This warning is heuristic.  For
\(\delta=0.05\), set
\begin{align}
 r_n&=\min\!\left\{1,\sqrt{\frac{\log(2Sn(n+1)/\delta)}{2n}}\right\},
 \nonumber\\
 L_{s,n}&=\max\{0,\widehat p_{s,n}-r_n\},\quad
 U_{s,n}=\min\{1,\widehat p_{s,n}+r_n\}.
 \label{eq:intervals}
\end{align}
A fresh evidence block supports separation when
\begin{equation}
 C_{s,n}=\mathbb I\!\left\{L_{s,n}>\tau,\quad
 L_{s,n}>\frac{1}{S-1}\sum_{j\neq s}U_{j,n}\right\}.
 \label{eq:certificate}
\end{equation}
A persistent request first latches only after \(C_{s,n}=1\) at two
consecutive fresh-evidence advances and the estimated remaining horizon is at
least two decisions.  These are nested prefixes, not independent replications.

\begin{figure*}[t]
\centering
\includegraphics[width=\textwidth]{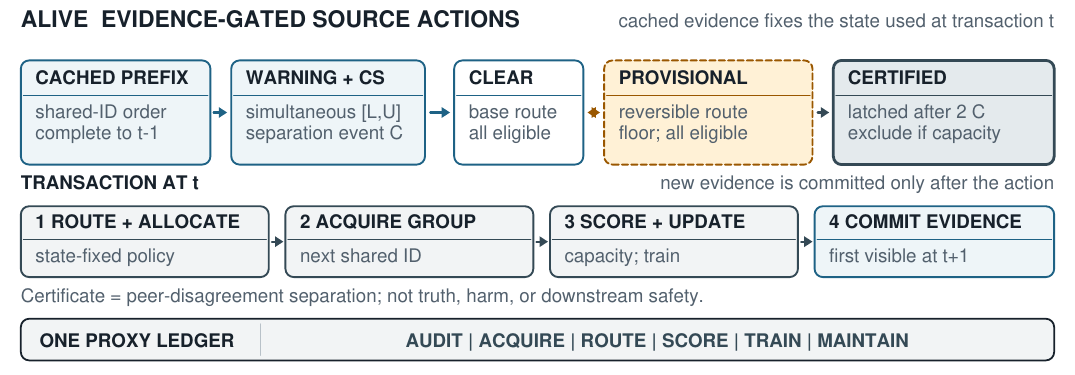}
\caption{One-decision evidence-to-action transaction.  State, routing,
allocation, exclusion request, and scoring eligibility are fixed from evidence
cached through \(t-1\).  New groups may update the audit data structure during
the transaction, but the frozen cache lets them affect actions only at \(t+1\).
Two certificate-positive advances latch a request; optimizer ineligibility is
activated only when remaining-source capacity can fill the batch.}
\label{fig:action-schematic}
\end{figure*}

Figure~\ref{fig:action-schematic} summarizes the state machine.  Warnings
change the next floor-bounded opportunity without latching or changing
eligibility; they do not undo past acquisition or training.  Two
certificate-positive advances latch a request, while each transaction activates
ineligibility only if remaining-source capacity fills the batch.  The
frozen 12.5/25\% audit shares, 15/40\% routing floor/cap, two confirmations, and
horizon guard are heuristics rather than theorem consequences.  Audit floors
preserve observability but do not recover a source; a utility certificate,
recovery rule, or value-of-information schedule is not part of the tested
method.

ALIVE--CBE (class-balanced entropy) composes the controller with its namesake
selector: a charged forward pass computes
\(h_t(x)=-\sum_k p_{\theta_t}(k\mid x)\log p_{\theta_t}(k\mid x)\), a
largest-remainder quota covers observed classes when feasible, and the highest
entropy eligible candidates are selected within class.  This selector is an
incumbent, not part of the novelty claim.

\subsection{Matched controller variants}

ALIVE--CBR (class-balanced random) changes only within-class ranking to a seeded
random permutation and
retains the controller, class quota, timing, and ledger.  Its horizon still uses
the counterfactual CBE step cost so selector choice cannot move certification,
while realized ranking cost is zero.  This is the matched strong-incumbent
composition, not a new selector.

The supplement specifies Full-Consensus and evidence- or clock-timed selector
switches as diagnostic boundaries and reports their complete outcomes.

\section{Evidence-to-Action Guarantee}
\label{sec:theory}

Condition on common support \(U\) and the complete potential-label table
\(Y=\{Y_s(i):s\in[S],i\in U\}\).  The comparable population is
\begin{align*}
 A&=\left\{i\in U:\max_y\sum_{s=1}^S
 \mathbb I\{Y_s(i)=y\}>S/2\right\},\\
 M&=|A|.
\end{align*}
The theorem idealizes the pseudorandom audit permutation as uniform and takes
probability over a hypothetical re-randomization conditional on the same
\((U,Y)\).  Domain-separating the audit seed from source-noise and optimizer
seeds prevents shared generator state, but is not itself a probabilistic
guarantee.  For \(M\geq1\), define the fixed finite-population target
\[
 p_s=M^{-1}\sum_{i\in A}\mathbb I\{Y_s(i)\neq m(i)\}.
\]
Let
\[
 \mathcal N=\left\{s:p_s\leq (S-1)^{-1}\sum_{j\neq s}p_j
 \ \text{or}\ p_s\leq\tau\right\}.
\]
Write \(\mathsf A_{\mathcal N}\), \(\mathsf L_{\mathcal N}\), and
\(\mathsf C_{\mathcal N}\) for any active exclusion, latched request,
and certificate-positive prefix, respectively, among sources in \(\mathcal N\).
The proof separates the evidence engine from the controller through
\begin{equation}
 \mathsf A_{\mathcal N}\subseteq\mathsf L_{\mathcal N}
 \subseteq\mathsf C_{\mathcal N}\subseteq\mathcal E_\delta^c.
 \label{eq:inheritance-chain}
\end{equation}
The action contract supplies the first two inclusions.  The base result below
supplies the last one with alpha-spent Hoeffding intervals; the PPR
specialization later gives a separate level-\(\delta\) engine under stronger
population conditions.

\paragraph{Theorem 1 (randomized-prefix coverage).}
If \(\pi\mid(U,Y)\) is a uniform permutation, then the intervals in
Equation~\eqref{eq:intervals} satisfy
\begin{equation}
 \Pr_\pi\!\left\{p_s\in[L_{s,n},U_{s,n}],\
 s\in[S],\ n\in[M]\mid U,Y\right\}\geq1-\delta .
 \label{eq:coverage}
\end{equation}
Denote the simultaneous-coverage event inside the braces by \(\mathcal E_\delta\).

\paragraph{Proof sketch.}
Restricting a uniform permutation of \(U\) to the fixed subset \(A\) gives a
uniform permutation of \(A\).  For fixed \((s,n)\), the first \(n\) binary
disagreements are sampled without replacement from a fixed Bernoulli
population.  Hoeffding's comparison gives
\(\Pr_\pi\{|\widehat p_{s,n}-p_s|\geq\epsilon\mid U,Y\}
\leq2e^{-2n\epsilon^2}\).  When the unclipped radius is below one,
substitution bounds failure by \(\delta/[Sn(n+1)]\); otherwise the clipped
interval is \([0,1]\).  Union bounding over sources and prefixes uses
\(\sum_{n\geq1}[n(n+1)]^{-1}=1\).  Simultaneity in \(n\) permits the prefix
length and decision time to depend on the observed prefix without another
optional-stopping correction. \(\square\)

Theorem~1 is an exact-real statement; the base controller uses ordinary binary
floating point without outward rounding and is not bit-level verified.  In the
synthetic PPR study, membership in the conservative weak-boundary PPR set is
exact-integer, but outward-padded floating hull endpoints enter the inherited
predicate.  Natural PPR instead evaluates its action-defining count
inequalities exactly in integers.

\paragraph{Corollary 1 (false action-predicate certification).}
On \(\mathcal E_\delta\) the two null branches are explicit.  If
\(p_s\leq\tau\), then \(L_{s,n}\leq p_s\leq\tau\); if
\(p_s\leq(S-1)^{-1}\sum_{j\ne s}p_j\), then
\(L_{s,n}\leq p_s\leq(S-1)^{-1}\sum_{j\ne s}p_j\leq
(S-1)^{-1}\sum_{j\ne s}U_{j,n}\).  Either branch contradicts one certificate
inequality.  Therefore
\begin{equation}
 \Pr_\pi\{\exists s\in\mathcal N\text{ ever certified}\mid U,Y\}
 \leq\delta .
 \label{eq:falsecert}
\end{equation}
The two-confirmation and horizon filters can only remove latch and action events
relative to certificate-positive prefixes; the confirmations use nested
prefixes and do not change \(\delta\) to \(\delta^2\).  If \(M=0\), no
disagreement mean or certification event is defined and the implementation
retains \([0,1]\).

\paragraph{Action-inheritance corollary (false persistent actions).}
Let a predictable controller choose warning actions, routing, and the number of
next prefix groups from evidence available before each decision.  Provided it
never skips, repeats, or replaces an identity according to labels, and every
hard action on source \(s\) requires a latch whose first entry followed a
certificate-positive prefix, the event of ever acting is contained in the event
of ever certifying.  Equation~\eqref{eq:falsecert} and
Equation~\eqref{eq:inheritance-chain} therefore bound by \(\delta\) the
probability of any hard action on \(s\in\mathcal N\).  Capacity fallback can
only remove active-exclusion events; it cannot create one without a latch.
The guarantee is consequently modular to non-latching policy changes that
preserve the randomized prefix.  It neither covers warning actions nor licenses
a label-dependent audit rule.

\paragraph{Corollary 2 (sufficient detection at a covered prefix).}
\label{cor:sufficient-detection}
For
\(\Delta_{\rm abs}=p_s-\tau\) and
\(\Delta_{\rm rel}=p_s-(S-1)^{-1}\sum_{j\ne s}p_j\), simultaneous coverage
implies that Equation~\eqref{eq:certificate} holds whenever both gaps are
positive and
\[
 r_n<\min\{\Delta_{\rm abs}/2,\Delta_{\rm rel}/4\}.
\]
Indeed, \(L_{s,n}\geq p_s-2r_n\) and
\(U_{j,n}\leq p_j+2r_n\); the supplement gives the complete argument.
Write \(g_s=\min\{\Delta_{\rm abs}/2,\Delta_{\rm rel}/4\}\) and
\(n_s^\star=\min\{n\in[M]:r_n<g_s\}\).  With \(A=U\), every
positive transaction advance followed by a fresh evaluation, predictable
per-transaction prefix increments at most \(b\), a following fresh evaluation,
and first-entry horizon at least two, Corollary~3 gives
\(n_{\mathrm{latch},s}\leq n_s^\star+2b\), using at most
\(S(n_s^\star+2b)\) cached action-defining audit labels; capacity may still
suppress activation.  This exact-real, schedule-conditional bound excludes all
other ledger costs and does not ensure utility.  If \(n_s^\star\) does not
exist by \(M\), the theorem supplies error control but no detection-power
claim.  The audit requirement can be substantial: for \(S=4\) and \(\delta=0.05\),
\(r_n\) is 0.149 at \(n=384\), 0.097 at \(n=1000\), and 0.034 at
\(n=10{,}000\).  The Hoeffding-spending radius also remains nonzero at a full
census, so this is a transparent conservative bound rather than a tight delay
characterization.  Under the nominal constructions, the sufficient boundary
first occurs at \(n=1{,}782\), 731, and 196 of \(M=10{,}000\) common identities
for e40, e60, and e80, respectively; e20 has
\(\Delta_{\rm abs}<0\), so no prefix can satisfy it.

\paragraph{Direct-contrast alternative.}
The supplement derives a strict-majority direct Hoeffding certificate that
pathwise contains the evaluated coordinatewise certificate, including under
clipping.  It remains unevaluated, its Bluebirds census bound is still below
zero.  It is a separate level-\(\delta\) procedure and is not unioned with
either reported certificate.

\paragraph{Known-size PPR specialization under strict majorities.}
Suppose all \(N=|U|=M\) identities are known in advance to have strict
majorities.  For \(K_s=Np_s\), the observed count \(x_{s,n}\), hypergeometric
likelihood \(\ell_{s,n}(k)\), uniform prior \(\pi_0(k)=1/(N+1)\), and posterior
\(\pi_{s,n}\), define
\[
 \mathcal C_{s,n}=\{k:\pi_0(k)/\pi_{s,n}(k)<S/\delta\}.
\]
Here and below, the ratio uses the extended-real convention
\(\pi_0(k)/0=+\infty\).
The posterior-ratio process is a nonnegative test supermartingale; Ville's
inequality \cite{waudbysmith2020confidence} and a source union give
\(\Pr_\pi\{K_s\in\mathcal C_{s,n},\forall s,n\mid U,Y\}\geq1-\delta\).
On feasible counts, the ratio is \(1/[(n+1)\ell_{s,n}(k)]\); the hull of
\(\mathcal C_{s,n}/N\) supplies anytime intervals and is exact at census.
This is an alternative level-\(\delta\) procedure: combining it with a
non-nested engine requires a pre-fixed error split.  Replacing unknown \(M\) by \(|U|\)
when any identity lacks a strict majority is invalid.  Synthetic and Bluebirds
replays used exact-integer membership in the conservative weak-boundary PPR set
at \(N=M=10{,}000\) and 108, respectively, with a strict majority at every
identity.  The fixed-panel replay is exploratory rather than a new confirmatory
family.

\paragraph{Exploratory finite-population correction (FPC).}
The base radius does not vanish at a census.  After the base certificate closed
no natural-panel outlier path, we separately evaluated a frozen,
non-confirmatory Serfling/census sensitivity.  At an ordinary comparable prefix (before support exhaustion), it
replaces
\(r_n\) by
\[
 \begin{aligned}
 \rho_{n,U}&=1-\frac{n-1}{|U|},\\
 r_n^{\rm FPC}&=\min\!\left\{1,
 \sqrt{\frac{\rho_{n,U}\log(2Sn(n+1)/\delta)}{2n}}\right\}.
 \end{aligned}
\]
If the unknown comparable population has size \(M\leq|U|\), the Serfling factor
\(1-(n-1)/M\) is no larger, so using \(|U|\) is conservative
\cite{serfling1974probability,bardenet2015concentration}; the same alpha-spending
argument applies.  Once every identity in \(U\) is exhausted, all comparable
values are known and the sensitivity closes with \(L_s=U_s=p_s\), making a false
population-outlier closure impossible.  Ordinary requests still need two fresh
prefixes and horizon; a census closure is statistical evidence for a future
capacity decision, not an activation.  This separately labeled sensitivity was not
used in the CIFAR trajectories and is not confirmatory evidence.

For external truth \(y^\star(i)\), let \(q_s\) and \(w\) be, respectively, the
source and strict-majority error rates restricted to \(A\).  Then
\(|p_s-q_s|\leq w\).  Thus translation
to source accuracy requires an additional bound on wrong majorities.  The
theorem says nothing about warnings, utility, generalization, or whether
exclusion is beneficial.

In the ordered-risk experiments, at least three construction-clean sources
make every retained majority correct, so \(w=0\) and \(p_s=q_s\) inside that
simulator.  In the rotating exact null, all four values satisfy \(p_s=1/4\),
so Equation~\eqref{eq:falsecert} applies to any source.  These consequences
depend on the constructions; they do not turn majority disagreement into a
general truth or deployment guarantee.  The evaluated bound is deliberately
loose and is not silently replaced.  The FPC diagnostic above
tests one classical correction under a separate label; other general,
without-replacement confidence sequences that incorporate identities without a
strict majority remain untested.

\section{Experimental Protocol}
\label{sec:protocol}

\paragraph{Data and environments.}
CIFAR-100 is used only for controller development; the reported
utility study uses a subsequent CIFAR-10 validation cache
\cite{krizhevsky2009learning}.  Both are encoded once by an ImageNet-pretrained
ResNet-18 \cite{he2016resnet,deng2009imagenet}, with split seed 2701.  Each run seed draws without replacement a 10,000-example pool from
the fixed 40,000-example training cache; paired methods share that pool and
all runs use the same fixed 10,000-example validation cache.  A one-hidden-
layer MLP of width 256 is trained with AdamW \cite{loshchilov2019decoupled}
(learning rate \(10^{-3}\), weight decay \(10^{-4}\)).  Candidate, maximum
batch, and minimum batch sizes are 512, 256, and 32.

The ordered-risk environments contain three construction-clean sources.  In
e20/e40/e80, one fourth source has a fixed 20\%, 40\%, or 80\% set of
identity-level label changes; repeated access to an identity returns the same
label.  In e60, two of five sources have independently fixed 60\% change sets.
The earlier development stage also contains an e80 low-overlap diagnostic,
reported in the supplement.  These are programmatic stress tests rather than
natural annotator data.  The exact null assigns one wrong label per
identity and rotates that role evenly across four sources, producing a unique
3--1 majority and exactly \(p_s=1/4\) for every source.

Budgets are 5, 10, 15, and 20\% of a fixed anchor.  The main ALIVE--CBE/CBR
comparisons use paired seeds 40--49; the known-size PPR study, which has a
strict majority at every identity, uses disjoint seeds 60--69.  Paired methods
share pools, budgets, and environment
definitions.  Methods, grids, endpoints, and consumers were fixed stagewise
before their corresponding outcomes; later variants remain validation-adaptive,
not independent confirmation.  The fixed validation cache supports conditional
comparisons, and the split-specific official test cache remains unopened.
Detailed protocols, complete results, study sequence, and provenance controls are in the supplement.

\paragraph{Natural complete-panel audit.}
Bluebirds contains 108 binary tasks fully labeled by 39 workers (4,212 labels)
\cite{welinder2010crowds}.  The base audit replays 10,000 fixed-seed
pseudorandom permutations at \((\delta,\tau)=(0.05,1/S)\), with two fresh
confirmations and an empirical comparator.  Ground truth enters only offline
majority-vote and pre-specified Dawid--Skene comparisons
\cite{dawid1979observer}.  FPC and exact-PPR replays were designed after their
parent natural-panel outcomes and frozen before their own outcomes.  They are
exploratory same-panel timing analyses; the panel has no features, routing,
capacity, or training utility.

\paragraph{Comparisons and inference.}
Strict consumers require complete finite grids, finite metrics, no overrun, and
exact reconstruction of prefix, state, routing, eligibility, fallback, and the
proxy ledger.  ALIVE--CBR holds CBR and controller timing fixed: routing-only is
the matched action ablation, while standalone CBR is the predeclared full-system
reference.  Empirical hard action is an uncalibrated diagnostic, not a same-FWER
baseline.  Other controller compositions and aggregation boundaries are
reported in the supplement.

Accuracy is primary and macro-F1 diagnostic; normalized area under the budget
curve (AUBC) integrates the four budgets.  Effects are averaged across
environments within each of ten paired seed clusters, not across 40 environment
sign-enumeration reference values require unestablished joint
sign-exchangeability \cite{ernst2004permutation}; Holm adjusts each pre-fixed
family \cite{holm1979simple}.  Paired intervals are descriptive, and
non-rejection is not equivalence.

\section{Results}
\label{sec:results}

\paragraph{Contract conformance and null restraint.}
On 40 repeated ALIVE--CBE exact-null cells, there were zero certificates,
requests, and active exclusions; the uncalibrated empirical controller acted
in all 40.  These cells share one construction and are a mechanism diagnostic,
not 40 independent confirmations of the nominal FWER.  Across the audited
ALIVE--CBE and ALIVE--CBR grids, strict replay found no certificate reopening,
eligibility violation, prefix skip or duplicate, label-selected identity, or
budget overrun.  Every action used only evidence cached through the preceding
decision.

In the ordered-risk ALIVE--CBR grid, all 40 cells in each of e40 and e80
latched and activated exactly the designated source, and all 40 e60 cells did so
for both designated sources.  All 40 e20 cells abstained from latching and active
exclusion, although 16/40 issued at least one non-latching warning.  Because the
e20 designated source has \(p_s=0.20<\tau=0.25\), this is the pre-fixed target
boundary rather than a low-risk power claim.

\begin{table}[t]
\centering
\small
\setlength{\tabcolsep}{3.2pt}
\begin{tabular}{@{}lcccc@{}}
\toprule
Env. & Warning & Latched request & Active & Median $t_{\rm act}$ \\
\midrule
e20 & 16/40 & 0/40 & 0/40 & --- \\
e40 & 40/40 & 40/40 & 40/40 & 12.5 \\
e60 & 40/40 & 40/40 & 40/40 & 7 \\
e80 & 40/40 & 40/40 & 40/40 & 3 \\
\bottomrule
\end{tabular}
\caption{ALIVE--CBR action funnel.  Entries before the median count runs in
which the event occurred at least once (40 per environment).  The implementation
records certification and latch together after the required two
certificate-positive advances; every latched request activated, with zero
capacity-blocked transaction steps.  Across 160 runs, the controller spent
796 transaction-steps in the non-latching
\textsc{Provisional} state.  The 160 latched
source paths remained active for a median 87
transactions (range 24--151).
Complete-group audit occupied 1,021,396/7,884,800
(13.0\%) acquired candidate slots.  $t_{\rm act}$ is the zero-based
first active-exclusion decision; ``---'' denotes no activation.}
\label{tab:action-funnel}
\end{table}

\begin{table*}[t]
\centering
\small
\setlength{\tabcolsep}{3.4pt}
\begin{tabular}{@{}p{0.25\textwidth}p{0.48\textwidth}p{0.22\textwidth}@{}}
\toprule
Reader question & Evidence-bound answer & Statistical role \\
\midrule
Does exact-null evidence trigger unsupported active exclusion?
  & ALIVE--CBE: \(0/40\);
    empirical immediate: \(40/40\)
  & Repeated mechanism diagnostic; not independent FWER replication \\
How does PPR evidence delay compare with Hoeffding?
  & Median separation prefix, PPR / Hoeffding:
    e40 \(96/304\),
    e60 \(62/171\),
    e80 \(48/48\)
  & Pre-known \(N=M\); every identity has a strict majority; no later on
    120/120 curves \\
What is the matched persistent-action increment?
  & ALIVE--CBR $-$ routing-only:
    \(+0.1935\;[+0.1172,\,+0.2697]\),
    10/10 seed wins
  & Met predeclared conditional reference criterion,
    \(p=.005859\) \\
Does the full system outperform standalone CBR?
  & ALIVE--CBR $-$ CBR:
    \(+0.1954\;[+0.0033,\,+0.3874]\),
    7/10 seed wins
  & Positive mean; system robustness criterion unmet
    (conditional ref. \(p=.097656\)) \\
\bottomrule
\end{tabular}
\caption{Evidence for the main CIFAR-10 validation questions.  The null
mechanism diagnostic uses ALIVE--CBE; the disjoint-seed delay comparison uses
separate level-0.05 PPR and Hoeffding engines for the same action target.  Each
identity count is one complete all-source audit group, not total label-query or
monetary cost.  Utility contrasts use ALIVE--CBR.  Effects are method minus
comparator in accuracy AUBC
(pp); brackets are descriptive paired-$t$ 95\% intervals, and wins count
positive differences over ten paired seed clusters.  The audit-only
decomposition and complete secondary results are in the supplement.  The
exact-null cells are repeated diagnostics; the official test remained unopened.}
\label{tab:lineage}
\end{table*}

\paragraph{Evidence-engine efficiency.}
The disjoint-seed PPR study completed 320 utility and 40 null runs with all
integrity and action-timing checks satisfied.  On the 120 high-risk curves, PPR
separated no later than Hoeffding on every curve and earlier on 100.  Median PPR
versus Hoeffding evidence counts were 96 versus 304 identities in e40, 62 versus
171 in e60, and 48 versus 48 in e80; all 40 null curves remained action-free.
Thus the tighter published engine materially reduces delay in some
margin--support regimes without claiming universal improvement.

Earlier evidence did not guarantee uniform downstream value.  PPR full minus
routing-only averaged \(+0.2235\) accuracy-AUBC pp, while e40 was
\(-0.0555\) pp; the predeclared all-environment utility criterion therefore
remained unmet.  The mechanism gain and downstream robustness are distinct
findings.

\paragraph{Matched value of persistent action.}
Within the frozen ALIVE--CBR controller, adding the persistent-action layer to
routing-only increased accuracy AUBC by \(+0.1935\) pp, with descriptive 95\%
interval \([+0.1172,+0.2697]\), positive differences in all ten seed clusters,
and conditional Holm reference \(p=0.005859\).  This matched contrast estimates
the persistent-action increment while holding audit, routing,
selector, timing, and pools fixed; it does not establish superiority to a
separate learning system.

A post-outcome descriptive decomposition clarifies the shared-budget mechanism.
Audit-only minus CBR was \(-0.2274\) pp, and routing-only minus audit-only was
\(+0.2293\) pp: non-latching routing approximately repaid the audit opportunity
cost before persistent action supplied the \(+0.1935\)-point increment.  These
contrasts are outside Holm and do not revise the predeclared endpoint.

\paragraph{Full-system operating regimes.}
ALIVE--CBR minus standalone CBR was \(+0.1954\) pp, with descriptive 95\%
interval \([+0.0033,+0.3874]\) and positive differences in 7/10 seed clusters.
The point estimate is positive, but the contrast did not meet the predeclared
multiplicity-adjusted robustness criterion (conditional Holm reference
\(p=0.097656\)).

\begin{table}[t]
\centering
\small
\setlength{\tabcolsep}{2.4pt}
\begin{tabular}{@{}lrrrrr@{}}
\toprule
Env. & 5\% & 10\% & 15\% & 20\% & AUBC \\
\midrule
e20 & \(-0.523\) & \(-0.230\) & \(+0.090\) & \(-0.051\) & \(-0.142\) \\
e40 & \(-0.693\) & \(-0.226\) & \(-0.111\) & \(-0.085\) & \(-0.242\) \\
e60 & \(+0.323\) & \(+0.525\) & \(+0.787\) & \(+0.778\) & \(+0.621\) \\
e80 & \(+0.210\) & \(+0.514\) & \(+0.683\) & \(+0.666\) & \(+0.545\) \\
\bottomrule
\end{tabular}
\caption{Environment--budget mean paired accuracy differences (ALIVE--CBR minus CBR, pp; seeds 40--49).  AUBC is the normalized trapezoidal integral across budgets; positive values favor ALIVE--CBR.}
\label{tab:r17-env-budget}
\end{table}

The aggregate separates into two regimes.  Accuracy AUBC was negative in e20
and e40 (\(-0.142\) and \(-0.242\) pp) but positive in e60 and e80
(\(+0.621\) and \(+0.545\) pp).  At 5\% budget, e20/e40 lost 0.523/0.693 pp.
The persistent-action layer can add value inside the matched controller, yet
audit cost and operating regime determine whether the full system improves on
CBR.  Complete CBE, predecessor, selector-switch, and aggregation results remain
in the supplement.

\paragraph{Natural-panel audit cost.}
The fixed Bluebirds population has 16 relative-disagreement outliers and 23
null workers.  Base ALIVE closed none of 160,000 outlier-worker--replay paths;
the empirical immediate rule closed every outlier path but requested at least
one null worker in every replay.  Majority-vote truth accuracy was 0.759 versus
0.889 for Dawid--Skene, underscoring that disagreement and truth are different
targets.

The exploratory Serfling procedure closed 96,117/160,000 paths before census
and 63,883 at census, with median prefix 105.  Separately level-0.05 PPR closed
128,917 before census and 31,083 at census, with median prefix
\(95/108=88.0\%\), or 3,705/4,212 complete-panel labels exposed.  Both had zero
null closures; PPR was no later on all paired paths and earlier on 80.6\%.
These are panel-exposure counts rather than query savings, and the replay has no
downstream utility endpoint.

\section{Discussion and Limitations}
\label{sec:limitations}

The three validity questions introduced in Section~\ref{sec:introduction}
resolve differently.  The randomized-prefix theorem addresses statistical
validity under its finite-population assumptions.  Semantic validity does not
follow: the focal source participates in the majority, a correct minority
expert can be an outlier, and correlated majority error can be invisible.
Decision validity also remains empirical.  The matched increment is positive
here, but the full-system regimes show why certificate validity alone cannot
answer it.

Four limitations bound the evidence.  First, the guarantee assumes fixed
support and labels, an ideal uniform permutation of the complete intersection,
and at least one comparable identity; partial overlap, drift, changing sources,
and call-dependent labels are outside scope.  The base construction abstains on
identities without a strict majority and excludes them from its estimand,
whereas the evaluated PPR specialization requires known \(N=M\) and a strict
majority at every identity.  Second, \(\tau=1/S\) and the
zero-margin relative predicate are policy choices, not a utility-derived
exclusion rule; no recovery or utility certificate is implemented.  Third,
downstream evidence uses synthetic label changes, a reused CIFAR validation
cache, ten seed clusters, and no published same-target, same-FWER action
baseline.  Bluebirds is one small complete panel without routing or training.
Finally, the ledger is a reproducible proxy rather than wall-clock, FLOPs,
energy, or monetary cost.  Independent natural downstream evaluation, resource
sensitivity, partial-overlap certificates, and drift-aware recovery remain open.

\section{Conclusion}

ALIVE treats persistent source action as a controlled interface rather than a
side effect of a warning score.  Its randomized prefix, evidence cache, latch,
and capacity check let predictable routers inherit a defined familywise
false-action bound.  In the frozen study, a published PPR engine sharply reduced
evidence delay in e40/e60, and persistent action added a positive matched
increment over routing-only.  The full-system endpoint and natural-panel cost
show that net utility still depends on margin, audit cost, and budget regime.
The resulting contract provides an auditable basis for designing and evaluating
persistent source actions.\label{page:last-technical-content}

\section*{Use of Generative AI}

GPT-5.6 Sol was used for language polishing. All AI-assisted revisions were reviewed and verified by the authors, who take full responsibility for the content of this paper.

\bibliography{references}

\end{document}


\maketitle
\appendix

\section{Scope and Reader Map}
\label{sec:supp-scope}

ALIVE separates empirical warnings, latched exclusion requests, and
capacity-feasible active exclusions under one shared learning budget.  This
supplement is organized by scientific function.  Section~\ref{sec:supp-algorithm}
gives the method variants, exact transaction, and accounting rules;
Section~\ref{sec:supp-theory} gives the finite-population guarantees and
alternative evidence engines; and Section~\ref{sec:supp-natural} reports the
complete-panel stress test.  Sections~\ref{sec:supp-protocol} and
\ref{sec:supp-outcomes} specify the experimental estimands, decision criteria,
and complete results.  Resource limitations, study sequence, reproducibility,
and non-claims follow in Sections~\ref{sec:supp-resource}--
\ref{sec:supp-limitations}.

The statistical model requires \(S\geq3\), stable record keys
and labels, and a nonempty all-source identity intersection.  These are
guarantee preconditions, not runtime-verified facts: empty support disables
certification and retains the incumbent.  The method does not cover disjoint,
pairwise-only, or dynamically changing source sets.  No action-defining branch
consumes construction risk flags, environment IDs, validation outcomes, or test
metrics.

The statistical statement is deliberately narrow.  It controls false
certification of a source that fails either conjunct of the pre-fixed
absolute-and-relative strict-majority-disagreement predicate in a fixed finite
population under an ideal uniform audit permutation.  It does not
certify truth, corruption, harm, downstream safety, accuracy, utility, regret,
calibration, or generalization.  The non-latching \textsc{Provisional} state has
no confidence guarantee.  Its routing state can clear, but past acquisition and
training are not undone.  This separation is the design principle: empirical
warnings may alter non-latching routing, whereas only simultaneous confidence
separation may latch an exclusion request; active ineligibility remains
capacity-feasible per transaction.

\section{Complete Algorithms and Accounting}
\label{sec:supp-algorithm}

\subsection{Setting and method families}

Let the source set be \([S]\), the candidate-window size at zero-based decision
\(t\) be \(N_t\), and the requested optimizer batch size be \(K_t\).
Controller evidence used at decision \(t\) contains only complete audit groups
collected through \(t-1\).  The fixed common audit support is
\(U=\bigcap_{s=1}^{S}U_s\).

\begin{table*}[t]
\centering
\small
\begin{tabular}{@{}p{0.27\textwidth}p{0.18\textwidth}
                p{0.20\textwidth}p{0.24\textwidth}@{}}
\toprule
Method & Routing evidence & Exclusion evidence & Incumbent\\
\midrule
ALIVE--CBE & empirical warning & certificate, latched &
class-balanced entropy (CBE)\\
ALIVE--CBE routing-only & empirical warning & never & CBE\\
Empirical quarantine (CBE) & empirical warning & warning, non-latching & CBE\\
ALIVE--CBR & empirical warning & certificate, latched &
class-balanced random (CBR)\\
ALIVE--CBR routing-only & empirical warning & never & CBR\\
ALIVE--PPR & empirical warning & exact uniform-PPR-hull certificate, latched & CBR\\
ALIVE--PPR routing-only & empirical warning; identical shadow PPR certificate &
never & CBR\\
Audit-only diagnostic & shadow warning audit; action disabled & never & uniform CBR\\
Empirical quarantine (CBR) & empirical warning & warning, non-latching & CBR\\
Adaptive-Switch & empirical warning & certificate, latched & CBR until the first
certificate decision; CBE from the next decision onward\\
Fixed-Switch & empirical warning & certificate, latched & CBR for \(t<8\); CBE
for \(t\geq8\)\\
CBR baseline & none & never & CBR\\
Full-Consensus & none & never & strict-majority-label CBE\\
\bottomrule
\end{tabular}
\caption{Reader-facing method variants.  Ablations share their parent method's
audit, allocator, and incumbent.  The audit-only control is an outcome-informed exploratory diagnostic;
Full-Consensus is a separate aggregation boundary.}
\label{tab:supp-methods}
\end{table*}

\subsection{Evidence predicates and state machine}

For source \(s\), let \(\widehat p_{s,n}\) be its disagreement rate over the
first \(n\) comparable audit identities, and let
\([L_{s,n},U_{s,n}]\) be Equation~\eqref{eq:supp-intervals}.  Before audit
labels are read, the action designer fixes \(\tau\in(0,1)\); the reported
experiments use \(\tau=1/S\).  This defines the action target, not an estimated
corruption rate.  After every source
has at least eight comparable observations, define
\begin{align}
 W_{s,n}
 &=\mathbb I\!\left\{
 \widehat p_{s,n}>\tau,\quad
 \widehat p_{s,n}>
 \frac1{S-1}\sum_{j\ne s}\widehat p_{j,n}
 \right\}, \label{eq:supp-warning}\\
 C_{s,n}
 &=\mathbb I\!\left\{
 L_{s,n}>\frac1{S-1}\sum_{j\ne s}U_{j,n},\quad
 L_{s,n}>\tau
 \right\}. \label{eq:supp-certificate}
\end{align}
The peer comparator in Equation~\eqref{eq:supp-certificate} is the
\emph{mean} of peer upper bounds, not their maximum.  It excludes the focal
source, whereas the strict majority defining disagreement includes it.
Equation~\eqref{eq:supp-warning} is heuristic and has no confidence
interpretation.

\begin{table}[t]
\centering
\small
\begin{tabular}{@{}p{0.21\columnwidth}p{0.28\columnwidth}
                p{0.41\columnwidth}@{}}
\toprule
State & Entry & Action\\
\midrule
\textsc{Clear} & no warning or certificate & uniform acquisition; 12.5\%
audit; all candidates eligible\\
\textsc{Provisional} & active empirical warning & non-latching monitored
routing; 25\% audit if no source is certified; candidates remain eligible\\
\textsc{Certified} & two consecutive fresh certificate advances and estimated
remaining horizon \(\geq2\) & latched exclusion request and monitoring;
12.5\% audit; capacity-feasible activation\\
\bottomrule
\end{tabular}
\caption{Aggregate controller semantics.  Capacity fallback suppresses one
exclusion transaction but does not clear a certificate.}
\label{tab:supp-states}
\end{table}

Confirmation streaks advance or reset only when the comparable prefix grows.
A source enters \textsc{Certified} after two consecutive fresh
\(C_{s,n}=1\) advances and only when the estimated remaining decision horizon
is at least two; certification is then latched for the stationary run.  A
not-yet-certified source with \(W_{s,n}=1\) is \textsc{Provisional}; otherwise
it is \textsc{Clear}.  The aggregate state is certified if any source is
latched, provisional if none is latched but at least one warning is active,
and clear otherwise.  Nested confirmations are not independent trials and do
not change \(\delta\) to \(\delta^2\).

\subsection{Exact per-decision procedure}

For ALIVE--CBE and ALIVE--CBR, one transaction is:
\begin{enumerate}
  \item Read the cached comparable prefix and compute
  \(\widehat p_{s,n}\), \(W_{s,n}\), \([L_{s,n},U_{s,n}]\), and \(C_{s,n}\).
  Advance/reset only fresh confirmation streaks; retain all latches.  This
  freezes every action-defining state for the transaction.
  \item Request audit fraction \(0.25\) exactly when at least one source is
  provisional and none is certified; otherwise request \(0.125\).  Compute the
  floor/ceiling-feasible integer source allocation from the cached state.
  \item Acquire the candidate window, placing next-unused audit identities into
  complete all-source groups and filling residual slots by ordinary source
  sampling.  Completed groups are appended to the audit data structure, but the
  frozen cache prevents them from changing any action in this transaction; they
  can first affect decision \(t+1\).
  \item Score all acquired candidates.  ALIVE--CBE performs one charged model
  forward and uses predictive entropy.  ALIVE--CBR draws one deterministic
  random permutation from the run generator, maps it to strictly descending
  scores, and charges no ranking forward.
  \item Use the cached state to form optimizer eligibility.  ALIVE requests
  exclusion only for certified sources, the empirical-hard-action ablation uses
  current warnings, and routing-only never excludes.  Activate a request only
  if the remaining candidates can fill \(K_t\); otherwise log capacity fallback
  and admit all candidates without clearing a latch.
  \item Compute class quotas, restrict to eligible candidates, select within
  class by descending score, train one update, and charge every acquisition,
  scoring, maintenance, and selected training example.
  \item Log the audit-permutation prefix, candidate and selected indices,
  source IDs, states, warning/certificate values, requested/realized audit,
  integer allocations, eligibility, fallback, ranking phase, and ledger events.
\end{enumerate}

The controller's first-entry horizon uses the counterfactual ALIVE--CBE
per-step denominator including an entropy-forward charge even for ALIVE--CBR
and pre-switch Adaptive-Switch.  This keeps the incumbent from changing the
certificate-entry rule.  The actual ledger nevertheless charges only executed
work.

\subsection{Exact-integer PPR procedure}

ALIVE--PPR retains the ALIVE--CBR incumbent, warning and routing policy,
uniform without-replacement audit prefix, two fresh-evidence confirmations,
horizon guard, monitoring, capacity fallback, ledger, and training protocol.
It replaces only the action-defining confidence endpoints by conservative
weak-boundary enlargements of the PPR hulls in Theorem~3.  Preflight requires a
known \(N=M=10{,}000\), all-source complete support, and a strict majority at
every identity.  Given \(\delta=a/b\), membership of count \(k\) is computed by
the exact integer comparison
\[
 a\binom Nn\leq bS(n+1)
 \binom{k}{x_{s,n}}\binom{N-k}{n-x_{s,n}}.
\]
Retaining equality is weakly more conservative than the theorem's strict set.
Hull endpoints are divided by \(N\) and padded outward before the inherited
strict floating-point certificate predicate.

The full method uses the resulting latched certificate for hard-action
eligibility; routing-only computes and latches the identical certificate but
sets the hard-action policy to \code{never}.  Paired paths must therefore be
identical until the first certificate and may diverge afterward only through
eligibility, selection, and downstream model quantities.  The reported
fresh-seed PPR replication made no controller or statistical change relative
to the initial PPR evaluation; it corrected only the analyzer--comparator
interface described in Section~\ref{sec:supp-study-sequence}.

\subsection{Class quotas and CBR ranking}

Let \(E_t\) be eligible candidate indices, \(\mathcal C_t\) their observed
classes, \(n_c\) class capacity, and \(K=\min(K_t,|E_t|)\).  If \(K\leq0\) or
no class is present, selection is empty.  If \(K<|\mathcal C_t|\), the \(K\)
largest observed classes receive one slot.  Otherwise each observed class
receives one slot; the residual is distributed proportionally to \(n_c-1\),
floored, then completed by largest fractional remainder without exceeding
capacity.

ALIVE--CBE ranks within class by entropy.  ALIVE--CBR draws
\(\operatorname{randperm}(|W_t|)\), maps it to strictly descending scores
\(1,1-1/|W_t|,\ldots,1/|W_t|\), restricts to \(E_t\), and applies the same
quota.  The fixed class and tie order is part of replay.

\subsection{Capped-simplex source allocation}

Without an actioned source, integer allocation is uniform and remainder slots
follow the fixed source order.  Otherwise, each provisional or certified source
receives a feasible lower monitoring target \(f=0.15\), non-actioned sources
receive the residual, and no continuous share exceeds
\(c=\max(0.40,1/S)\).  Residual mass is redistributed through the capped
simplex; therefore \(15\%\) is neither an equality nor an upper bound for an
actioned source.

Shares are multiplied by \(N_t\), floored, and completed by largest remainder.
A repair enforces
\[
 k_{\min}=\left\lceil\min(0.15,1/S)N_t-10^{-12}\right\rceil
\]
by transfers from sources above \(k_{\min}\); if
\(S k_{\min}>N_t\), the pre-repair feasible allocation is retained.  For
\(N_t=512,S=4\), three monitored sources receive \((103,102,102)\) and the
other source 205, up to the fixed source order; all four monitored sources receive 128
each.  Largest-remainder integerization can exceed the continuous ceiling by
one candidate.

\subsection{Audit construction and quarantine}

On first use, a separate audit generator permutes \(U\) once.  With requested
fraction \(\alpha_t\),
\begin{align*}
 R_t&=\lfloor\alpha_tN_t\rfloor,
 &G_t&=\left\lfloor R_t/S\right\rfloor,\\
 g_t&=\min\{G_t,\min_s n_{st},|U|-\mathrm{cursor}_t\}.
\end{align*}
Exactly \(g_t\) identities are queried from every source, giving \(Sg_t\)
audit slots.  Shortfall separately records complete-group rounding,
allocation clipping, support exhaustion, and final-budget shrinkage.  Audit
slots occupy the ordinary source allocation and are charged once.  Only a
unique strict majority creates a comparable observation; identities without a
strict majority abstain.
The cursor cannot skip, revisit, or replace identities.  Exhaustion freezes
the prefix rather than substituting source-specific records.

For the confidence-gated methods, let \(Q_t\) contain candidates from
certified sources.  Exclusion is requested when \(Q_t\ne\varnothing\) and is
active only if \(|W_t\setminus Q_t|\geq K_t\).  Otherwise all candidates are
eligible and capacity fallback is logged.  A suppressed request is not an
active quarantine or false isolation.

The 12.5/25\% audit shares and 15/40\% monitoring floor/cap are fixed
heuristics.  After a latch, the monitoring floor and 12.5\% audit maintain
observability of additional source divergence and capacity state; they do not
recover the latched source.  Their continuing ledger charge is a declared
utility cost and a plausible contributor to the lack of a CBR win.  An
independently validated recovery/epoch reset or value-of-information audit
schedule is future work and was not tested.

\subsection{Adaptive and fixed ranking schedules}

Adaptive-Switch inherits the ALIVE--CBE audit, state, routing, horizon, monitoring, and quarantine
without change.  Before certification it uses CBR and has zero ranking
cheap-evaluation cost.  On the first decision with a nonempty certified-source
latch, quarantine may act immediately, but ranking is still CBR.  That
transaction irreversibly arms entropy.  At exactly the next decision, CBE
becomes active and the actual candidate-forward cost is reserved before
selection and charged once.  The switch never rolls back and has no
environment branch.

Fixed-Switch preserves the same controller and quarantine, but fixes ranking independently
of controller state: CBR for \(t=0,\ldots,7\) and CBE for every \(t\geq8\).
Decision 8 is reserved before selection.  Its clock was fixed from the pooled
zero-based median first-certificate decision 7 over 120 ALIVE--CBR high-risk
curves, plus the one-decision delay used by Adaptive-Switch.  No Adaptive-Switch
outcome contributed.

\subsection{Full-consensus boundary}

Full-Consensus is not a state-controller row.  Each decision samples exactly 512 identity
positions with replacement from the all-source intersection, retains
duplicates, and materializes the aligned proposal from every source.  It fails
closed unless aligned features are exactly equal.  All \(S\) labels are
queried; a target is retained only when one label count is strictly greater
than \(S/2\), with no fallback or resampling for an identity without a strict
majority.  Proposal
features are ranked by entropy, the same quota routine selects at most 256
examples, and training uses logical source \code{consensus}.  No feedback,
warning, audit, routing, certificate, allocation, or quarantine path executes.

Per full step, Full-Consensus charges \(S\times512\) source labels, 512 entropy scores,
\(S\times512\) majority-tally events, and the realized training count.  Nominal
full-step totals are 343.04 units for four sources and 358.40 for five.  The
last step is shrunk before selection and then charged at realized count.  The
synthetic construction has three clean sources, so strict majority equals
simulator truth; this is a strong clean-target boundary, not a natural-worker
truth-inference result.

\subsection{Ledger semantics}

\begin{table}[t]
\centering
\small
\begin{tabular}{@{}p{0.22\columnwidth}p{0.20\columnwidth}
                p{0.44\columnwidth}@{}}
\toprule
Stage & Unit cost & Count\\
\midrule
source acquisition & 0.02 & acquired candidate/source-label slots\\
cheap evaluation & 0.05 & actually scored candidates\\
fine evaluation & 0.20 & fine-scored candidates\\
training & 1.00 & selected examples\\
maintenance & 0.01 & declared maintenance/tally events\\
refresh & configured & realized refresh action\\
\bottomrule
\end{tabular}
\caption{Proxy-accounting coefficients used throughout the experiments.  These units are not FLOPs,
energy, currency, or universal latency.}
\label{tab:supp-costs}
\end{table}

Strict auditors reconstruct every ledger event and require realized cost not
to exceed the declared budget.  Acquisition/audit slots are not double
charged.  ALIVE--CBE, post-switch Adaptive-Switch, Fixed-Switch after decision 7, and
Full-Consensus charge entropy forwards; ALIVE--CBR, CBR, and pre-switch
Adaptive-Switch do not.  Trained-example exposure counts
selected indices, including repeat exposures.

\section{Finite-Population Audit Guarantee}
\label{sec:supp-theory}

\subsection{Probability space and estimand}

Let \(U\) be the finite common audit support and
\(Y=\{Y_s(i):s\in[S],i\in U\}\) the complete potential-label table.  All
results condition on realized \((U,Y)\).  The mathematical randomization is
\begin{equation}
 \pi\mid(U,Y)\sim\operatorname{Unif}(\mathfrak S_U).
 \label{eq:supp-randomization}
\end{equation}
No independence across sources or identities and no i.i.d.\ superpopulation
model are assumed.

The audit-seed lookup is generated separately from source-noise and training
generators.  Domain separation prevents accidental generator-state sharing;
it does not prove Equation~\eqref{eq:supp-randomization}.  Once a seed is
fixed, the realized permutation and all decisions are deterministic.  The
nominal probability refers to ideal rerandomization conditional on the same
\((U,Y)\), not residual randomness in a run or a frequency over ten seeds.

Define
\begin{align*}
 A&=\left\{i\in U:
 \max_y\sum_{s=1}^{S}\mathbb I\{Y_s(i)=y\}>S/2\right\},\\
 M&=|A|.
\end{align*}
For \(i\in A\), the strict-majority label \(m(i)\) is unique.  If \(M=0\),
the comparable-prefix set is empty and the implementation retains \([0,1]\)
intervals without advancing certification.  Assume \(M\geq1\), and define
\[
 Z_s(i)=\mathbb I\{Y_s(i)\neq m(i)\},\qquad
 p_s=M^{-1}\sum_{i\in A}Z_s(i).
\]

\paragraph{Lemma 1 (restriction of a random permutation).}
Conditional on \((U,Y)\), the relative order induced by \(\pi\) on fixed
subset \(A\) is uniform over the \(M!\) permutations of \(A\).

\paragraph{Proof.}
Fix an ordering of \(A\).  For each choice of the \(M\) positions occupied by
\(A\), exactly \((|U|-M)!\) permutations of the remaining identities induce
that ordering.  Summing over the \(\binom{|U|}{M}\) position choices gives the
same count for every ordering of \(A\). \(\square\)

\subsection{Simultaneous source-by-prefix coverage}

Let \(a_1,\ldots,a_M\) be the order induced on \(A\).  For \(s\in[S]\),
\(n\in[M]\), and \(\delta\in(0,1)\), define
\begin{align}
 \widehat p_{s,n}
 &=n^{-1}\sum_{k=1}^{n}Z_s(a_k),\nonumber\\
 \widetilde r_n
 &=\sqrt{\frac{\log(2S n(n+1)/\delta)}{2n}},
 \qquad r_n=\min\{1,\widetilde r_n\},\nonumber\\
 L_{s,n}&=\max\{0,\widehat p_{s,n}-r_n\},\nonumber\\
 U_{s,n}&=\min\{1,\widehat p_{s,n}+r_n\}.
 \label{eq:supp-intervals}
\end{align}

\paragraph{Theorem 1 (randomized audit coverage).}
Under Equation~\eqref{eq:supp-randomization},
\[
 \Pr_\pi\!\left\{
 \forall s\in[S],\ \forall n\in[M]:
 p_s\in[L_{s,n},U_{s,n}]
 \,\middle|\,U,Y
 \right\}\geq1-\delta .
\]

\paragraph{Proof.}
By Lemma~1, for fixed \(s,n\), the values
\(Z_s(a_1),\ldots,Z_s(a_n)\) form a simple random sample without replacement
from the fixed Bernoulli population \(\{Z_s(i):i\in A\}\).  Hoeffding's
comparison for sampling without replacement \cite{hoeffding1963probability}
gives
\[
 \Pr_\pi\!\left\{
 |\widehat p_{s,n}-p_s|\geq\epsilon
 \,\middle|\,U,Y\right\}\leq2\exp(-2n\epsilon^2).
\]
If \(\widetilde r_n<1\), substitution yields
\[
 2\exp\!\left[-\log\!\left(\frac{2S n(n+1)}{\delta}\right)\right]
 =\frac{\delta}{S n(n+1)}.
\]
If \(\widetilde r_n\geq1\), then \(r_n=1\) and clipping produces
\([0,1]\), with failure probability zero.  Let \(F\) denote any
source--prefix failure.  A union bound gives
\begin{align*}
 \Pr_\pi(F\mid U,Y)
 &\leq\sum_{s=1}^{S}\sum_{n=1}^{M}\frac{\delta}{S n(n+1)}\\
 &\leq\delta,
\end{align*}
because \(\sum_{n\geq1}[n(n+1)]^{-1}=1\). \(\square\)

\paragraph{Adaptive-prefix consequence.}
The implementation consumes one cached permutation and adds only complete
all-source groups.  It may use past evidence to choose a later audit request,
stop at a budget boundary, or stop at support exhaustion.  After majority
abstentions, observed comparable identities still form an initial segment of
\((a_1,\ldots,a_M)\).  The theorem is simultaneous in \(n\), so evaluating at
these adaptive times requires no additional optional-stopping correction.
Label-dependent identity selection, revisits, or call-dependent labels would
violate the premise.

This is Hoeffding's known comparison plus explicit alpha spending, not a new
confidence-sequence construction.  Tighter finite-population corrections and dedicated without-
replacement confidence sequences \cite{serfling1974probability,waudbysmith2020confidence}
are not used by the evaluated CIFAR controller; the separately labeled natural
diagnostic below applies the classical Serfling factor.

\subsection{Hard-action corollary}

Let
\[
 \overline U_{-s,n}=\frac1{S-1}\sum_{j\neq s}U_{j,n}.
\]

\paragraph{Corollary 1 (false action-predicate certification).}
On simultaneous coverage,
\[
 L_{s,n}>\overline U_{-s,n}
 \Longrightarrow
 p_s>\frac1{S-1}\sum_{j\neq s}p_j,
\]
and \(L_{s,n}>\tau\) implies \(p_s>\tau\).  Thus, for
\[
 \mathcal N=\left\{s:
 p_s\leq\frac1{S-1}\sum_{j\neq s}p_j
 \ \text{or}\ p_s\leq\tau\right\},
\]
\[
 \Pr_\pi\{\exists s\in\mathcal N\text{ ever entering
 \textsc{Certified}}\mid U,Y\}\leq\delta.
\]

\paragraph{Proof.}
On coverage,
\[
 p_s\geq L_{s,n}>
 \frac1{S-1}\sum_{j\neq s}U_{j,n}
 \geq\frac1{S-1}\sum_{j\neq s}p_j.
\]
For the first branch of \(\mathcal N\), certification is contained in the
complement of simultaneous coverage.  For the second branch,
\(L_{s,n}>\tau\) contradicts \(p_s\geq L_{s,n}\) and \(p_s\leq\tau\).
Taking their union proves the claim. \(\square\)

Any predictable warning, routing, or adaptive audit-size policy preserves this
bound when it consumes only the next groups of the same prefix and every hard
action requires a latch whose first entry followed a certificate-positive
prefix.  Event-wise, ever acting is then contained in ever certifying.  The
bound does not cover warning quality or label-dependent identity selection.

\paragraph{Corollary 2 (sufficient detection at a covered prefix).}
\label{cor:supp-sufficient-detection}
Suppose source \(s\) has positive gaps
\[
 \Delta_{\rm abs}=p_s-\tau,
 \qquad
 \Delta_{\rm rel}=p_s-\frac1{S-1}\sum_{j\ne s}p_j.
\]
On simultaneous coverage, the certificate predicate in
Equation~\eqref{eq:supp-certificate} holds at prefix \(n\) if
\[
 r_n<\min\!\left\{\frac{\Delta_{\rm abs}}2,
                       \frac{\Delta_{\rm rel}}4\right\}.
\]

\paragraph{Proof.}
Coverage gives
\(\widehat p_{s,n}\geq p_s-r_n\) and
\(\widehat p_{j,n}\leq p_j+r_n\).  Clipping can only strengthen the resulting
one-sided inequalities, so
\begin{align*}
 L_{s,n}&\geq p_s-2r_n,\\
 \frac1{S-1}\sum_{j\ne s}U_{j,n}
 &\leq\frac1{S-1}\sum_{j\ne s}p_j+2r_n.
\end{align*}
The absolute inequality follows from \(2r_n<\Delta_{\rm abs}\); subtracting
the peer bound from the focal lower bound leaves more than
\(\Delta_{\rm rel}-4r_n>0\).  Both certificate conjuncts therefore hold.
\(\square\)

\paragraph{Corollary 3 (schedule-conditional audit-to-latch label bound).}
\label{cor:supp-audit-to-latch}
Assume \(A=U\), and let
\(0=n_0<n_1<\cdots\) be the cached comparable-prefix sizes at fresh
evaluations.  Every positive transaction audit advance produces the next
fresh evaluation, and each such predictable increment is at most
\(b\in\mathbb N\).  For the gaps in Corollary~2, define
\begin{align*}
 g_s&=\min\{\Delta_{\rm abs}/2,\Delta_{\rm rel}/4\},\\
 n_s^\star&=\min\{n\in[M]:r_n<g_s\}.
\end{align*}
Suppose \(g_s>0\), \(n_s^\star\) exists, the schedule reaches a first
\(n_{k^\star}\geq n_s^\star\) and has a following fresh evaluation
\(n_{k^\star+1}\), and the first-entry horizon guard at that second
evaluation is at least two.  On the simultaneous-coverage event, source \(s\)
latches by
\[
 n_{\mathrm{latch},s}\leq n_s^\star+2b.
\]
At most \(S(n_s^\star+2b)\) cached action-defining audit labels are used
through the latch.  Including the new audit groups acquired after state is
frozen in the latch transaction gives at most \(S(n_s^\star+3b)\) audit
labels by the transaction's end.  If
\(|W_t\setminus Q_t|\geq K_t\), activation occurs in that same transaction;
otherwise the request remains latched but this result gives no activation-delay
bound.
The label counts exclude ordinary candidate acquisition, scoring, maintenance,
and training; they are neither wall-clock nor total-budget bounds.

\paragraph{Proof.}
Put \(c=2S/\delta\).  For positive real \(n\), the derivative of
\(\log(cn(n+1))/n\) has the sign of
\[
 1+\frac{n}{n+1}-\log(cn(n+1)),
\]
which is negative because \(S\geq3\), \(\delta\in(0,1)\), and
\(n\geq1\).  Hence \(r_n\) decreases.  By minimality of \(k^\star\)
and the increment bound,
\[
 n_{k^\star}\leq n_s^\star+b,
 \qquad n_{k^\star+1}\leq n_s^\star+2b.
\]
Both radii are below \(g_s\), so Corollary~2 makes both fresh evaluations
certificate-positive on simultaneous coverage.  The second advance completes
the two-fresh streak, and the assumed horizon guard latches the request.
Because \(A=U\), each prefix identity contributes one label from every
source, proving the cached count.  State is frozen before the transaction's
new acquisition, which contributes at most one further \(Sb\) block and
proves the post-transaction count.  The capacity statement follows from the
activation predicate.  Finally, activation is contained in latch, which is
contained in certification, so Corollary~1 still bounds false activation by
\(\delta\).  \(\square\)

The preceding results prove the main-paper detection statement and its
schedule consequence.  They are sufficient conditions, not necessary
sample-complexity bounds.
Two fresh-prefix confirmations and the horizon guard remain required before a
latch.  They only remove certification events, so they neither weaken
Corollary~1 nor square its error level.  Because a latched exclusion request
requires certification, active exclusion of a member of \(\mathcal N\) is also
contained in the same event.  The corollaries say nothing about utility,
safety, generalization, or whether exclusion improves learning.

\subsection{Derived alternative confidence constructions}
\label{sec:supp-derived-confidence}

The direct construction remains unevaluated.  The PPR construction was
preselected for the synthetic ALIVE--PPR trajectories reported below.  Each is
a complete alternative level-\(\delta\) certificate.  By the pathwise dominance
proved below, on any common source--prefix path the base certificate event is
contained in the direct certificate event.  Their pointwise union therefore
equals the direct certificate and incurs no additional multiplicity.  A union
with a non-nested engine, such as PPR, requires a pre-fixed error split.

\paragraph{Theorem 2 (strict-majority direct-contrast lower bounds).}
Put \(h=S-1\), \(q=\lfloor h/2\rfloor\), and
\begin{align*}
 D_s(i)&=Z_s(i)-h^{-1}\sum_{j\ne s}Z_j(i),\\
 d_s&=M^{-1}\sum_{i\in A}D_s(i)
     =p_s-h^{-1}\sum_{j\ne s}p_j,\\
 \widehat d_{s,n}&=n^{-1}\sum_{k=1}^nD_s(a_k),\qquad
 R=1+q/h,\\
 G_{s,n}&=\widehat d_{s,n}-Rr_n.
\end{align*}
Under Equation~\eqref{eq:supp-randomization},
\[
\begin{aligned}
 \Pr_\pi\!\bigl\{&L_{s,n}\leq p_s,\ G_{s,n}\leq d_s,\\
 &\forall s\in[S],\ \forall n\in[M]\mid U,Y\bigr\}\geq1-\delta.
\end{aligned}
\]
Consequently, the strict alternative certificate
\begin{equation}
 C^D_{s,n}=\mathbb I\{L_{s,n}>\tau,\ G_{s,n}>0\}
 \label{eq:supp-direct-certificate}
\end{equation}
controls at level \(\delta\) the probability of ever certifying any
\(s\in\mathcal N\).

\paragraph{Proof.}
A unique strict majority leaves at most \(q\) dissenting sources on each
comparable identity.  If source \(s\) agrees, then \(D_s(i)\geq-q/h\); if it
dissents, then \(0\leq D_s(i)\leq1\).  Hence every \(D_s(i)\in[-q/h,1]\), an
interval of width \(R\).  For \(\widetilde r_n<1\), one-sided
without-replacement Hoeffding bounds give
\begin{align*}
 \Pr_\pi\{\widehat p_{s,n}-p_s\geq r_n\mid U,Y\}
 &\leq \frac{\delta}{2Sn(n+1)},\\
 \Pr_\pi\{\widehat d_{s,n}-d_s\geq Rr_n\mid U,Y\}
 &\leq \frac{\delta}{2Sn(n+1)}.
\end{align*}
For \(\widetilde r_n\geq1\), \(r_n=1\) makes both lower bounds
pathwise valid: \(L_{s,n}=0\leq p_s\) and
\(G_{s,n}\leq1-R=-q/h\leq d_s\).  Summing the two failure probabilities
over sources and prefixes uses \(\sum_{n\geq1}[n(n+1)]^{-1}=1\).
On the resulting event, \(C^D_{s,n}=1\) strictly implies \(p_s>\tau\) and
\(d_s>0\), which excludes both branches of \(\mathcal N\). \(\square\)

\paragraph{Proposition (pathwise dominance).}
At the same \((\delta,\tau)\), every prefix satisfying the evaluated
coordinatewise certificate in Equation~\eqref{eq:supp-certificate} also
satisfies Equation~\eqref{eq:supp-direct-certificate}, including under
clipping.

\paragraph{Proof.}
For \(r_n=1\), the evaluated strict relative inequality is impossible because
\(L_{s,n}=0\) while peer upper bounds are nonnegative.  Suppose \(r_n<1\), and
write
\[
\begin{aligned}
 \bar p_{-s,n}&=h^{-1}\sum_{j\ne s}\widehat p_{j,n},\\
 c_j&=U_{j,n}-\widehat p_{j,n}\\
 &=\min\{r_n,1-\widehat p_{j,n}\}.
\end{aligned}
\]
The evaluated relative inequality has a nonnegative right side, so it implies
\(L_{s,n}=\widehat p_{s,n}-r_n\) and
\[
 \widehat d_{s,n}>r_n+h^{-1}\sum_{j\ne s}c_j.
\]
For \(r_n\in[0,1]\), \(c_j\geq r_n(1-\widehat p_{j,n})\).  Moreover, at
most \(q\) peers dissent on each identity, so \(\bar p_{-s,n}\leq q/h\).
Thus
\[
 h^{-1}\sum_{j\ne s}c_j
 \geq r_n(1-\bar p_{-s,n})
 \geq r_n(1-q/h)\geq r_nq/h,
\]
because \(q/h\leq1/2\).  Therefore
\(\widehat d_{s,n}>r_n(1+q/h)=Rr_n\).  The absolute conjunct
\(L_{s,n}>\tau\) is unchanged. \(\square\)

For the Bluebirds population, \(h=38\), \(q=19\), and
\(Rr_{108}=0.4174\), whereas the largest population relative margin is
0.3826.  Thus even the dominating direct certificate fails its relative
conjunct at census; it does not resolve the evaluated base method's zero-power
finding.

\paragraph{Theorem 3 (known-size uniform-PPR sets with strict majorities).}
Assume every identity in \(U\) is known in advance to have a strict majority,
so \(N=|U|=M\) is known.  For source \(s\), let \(K_s=Np_s\) and
\(x_{s,n}=\sum_{j=1}^n Z_s(a_j)\).  For \(k\in\{0,\ldots,N\}\), define the
hypergeometric likelihood
\[
 \ell_{s,n}(k)=
 \frac{\binom{k}{x_{s,n}}\binom{N-k}{n-x_{s,n}}}{\binom Nn},
\]
with infeasible binomial coefficients equal to zero.  Under the uniform prior
\(\pi_0(k)=1/(N+1)\), let
\begin{equation}
\begin{aligned}
 \pi_{s,n}(k)&=
 \frac{\pi_0(k)\ell_{s,n}(k)}
      {\sum_{u=0}^N\pi_0(u)\ell_{s,n}(u)},\\
 \mathcal C_{s,n}&=\{k:\pi_0(k)/\pi_{s,n}(k)<S/\delta\}.
\end{aligned}
 \label{eq:supp-ppr-set}
\end{equation}
Here and below, the ratio uses the extended-real convention
\(\pi_0(k)/0=+\infty\).
Then
\[
\begin{aligned}
 \Pr_\pi\!\bigl\{&K_s\in\mathcal C_{s,n},\ \forall s\in[S],\\
 &\forall n\in\{0,\ldots,N\}\mid U,Y\bigr\}\geq1-\delta.
\end{aligned}
\]
When \(\mathcal C_{s,n}\neq\varnothing\), its hull divided by \(N\) is an
anytime-valid interval for \(p_s\); an empty set triggers no certificate.
Replacing \([L_{s,n},U_{s,n}]\) in the original strict certificate by these
hulls therefore controls the same false-action family.

\paragraph{Proof.}
Fix \(s\) and its true count \(K_s=k\).  Let \(P_k\) denote the law of the
count history under that count and let \(Q=\sum_{u=0}^N\pi_0(u)P_u\) be the
mixture law.  Under the count filtration
\(\mathcal F^s_n=\sigma(x_{s,0},\ldots,x_{s,n})\), the ratio
\[
 E_{s,n}(k)=\frac{\pi_0(k)}{\pi_{s,n}(k)}
 =\frac{\sum_{u=0}^N\pi_0(u)\ell_{s,n}(u)}{\ell_{s,n}(k)}
\]
equals \(Q(h_n)/P_k(h_n)\) on every history \(h_n\) in the support of
\(P_k\).  It is a nonnegative test supermartingale initialized at one.  Indeed,
if \(\mathcal E_k(h_n)\) is the set of one-step extensions having positive
\(P_k\)-probability, then
\[
\begin{aligned}
 \mathbb E_{P_k}[E_{s,n+1}(k)\mid h_n]
 &=\frac{\sum_{h_{n+1}\in\mathcal E_k(h_n)}Q(h_{n+1})}{P_k(h_n)}\\
 &\leq \frac{Q(h_n)}{P_k(h_n)}=E_{s,n}(k).
\end{aligned}
\]
The inequality allows mixture mass on extensions that are impossible under
the true count \(k\); this support leakage is why equality need not hold.
At census, \(E_{s,N}(k)=\pi_0(k)=1/(N+1)\), consistent with a
supermartingale rather than a martingale in general.
Ville's inequality \cite{waudbysmith2020confidence} under the
random-permutation law gives
\(\Pr_\pi\{\exists n:E_{s,n}(k)\geq S/\delta\mid U,Y\}\leq\delta/S\).
The strict set boundary in Equation~\eqref{eq:supp-ppr-set} makes exclusion of
the true count exactly such a crossing; union bounding over \(S\) sources proves
the claim and requires no independence across sources.  Vandermonde's identity gives
\[
 \sum_{u=0}^N\binom{u}{x}\binom{N-u}{n-x}
 =\binom{N+1}{n+1},
\]
so for the uniform prior
\[
 \frac{\pi_0(k)}{\pi_{s,n}(k)}
 =\frac{1}{(n+1)\ell_{s,n}(k)}
\]
on feasible support.  If \(\delta=a/b\) in lowest terms, exact membership is
\[
\begin{aligned}
 k\in\mathcal C_{s,n}
 &\quad\Longleftrightarrow\quad
 a\binom Nn < bS(n+1)\\
 &\hspace{34mm}\times
 \binom{k}{x_{s,n}}\binom{N-k}{n-x_{s,n}}.
\end{aligned}
\]
The inequality is strict: equality is excluded by the definition of
\(\mathcal C_{s,n}\).  At \(n=N\), only \(k=x_{s,N}\) is feasible, hence the
hull is the exact singleton \(\{p_s\}\). \(\square\)

Posterior ratios may be evaluated descriptively in log space, with zero
likelihood mapped to infinity.  The action-defining ALIVE--PPR membership test
instead uses the exact-integer non-strict variant of the membership inequality
above, conservatively retaining equality.  Hull endpoints are converted to
binary floating point with 64-ULP
outward padding before the inherited certificate predicate; this narrows but
does not formally eliminate program-level arithmetic risk.  This specialization
applies to Bluebirds because all 108 tasks are known, observed, and have a
strict majority, and to the mechanically checked synthetic ALIVE--PPR panels
with \(N=M=10{,}000\).  It does not apply when any identity lacks a strict
majority, making \(M\) unknown: substituting \(|U|\) for \(M\) changes both the
support and likelihood and is
not a conservative shortcut.  Synthetic downstream PPR trajectories and the later exploratory Bluebirds
mechanism replay are reported below under separate claim boundaries.

\subsection{Majority identifiability}

Let \(y^\star(i)\) be an external true label on \(A\), and define
\begin{align*}
 q_s&=M^{-1}\sum_{i\in A}\mathbb I\{Y_s(i)\neq y^\star(i)\},\\
 w&=M^{-1}\sum_{i\in A}\mathbb I\{m(i)\neq y^\star(i)\}.
\end{align*}
Let
\[
 d_s=p_s-\frac1{S-1}\sum_{j\neq s}p_j,\qquad
 d_s^\star=q_s-\frac1{S-1}\sum_{j\neq s}q_j.
\]

\paragraph{Proposition 1 (disagreement versus true error).}
For every source, \(|p_s-q_s|\leq w\), and
\(|d_s-d_s^\star|\leq2w\).

\paragraph{Proof.}
When \(m(i)=y^\star(i)\), majority-disagreement and true-error indicators
coincide.  On each of the remaining \(wM\) identities, their difference has
absolute value at most one, giving \(|p_s-q_s|\leq w\).  Apply this bound to
the focal source and peer mean, then use the triangle inequality. \(\square\)

Thus \(w=0\) permits a conditional true-label interpretation on \(A\);
without a bound on \(w\), it does not.  Unanimous common-mode error is
invisible, a correct minority can be the unique majority-disagreement outlier,
and tied identities are excluded from the estimand.

\subsection{Construction-specific consequences}

Every ordered-risk environment contains three construction-clean sources
among four or five sources.  On common support they form a correct strict
majority, so \(A=U,w=0\), and clean sources have \(p_s=0\).  Within a fixed
synthetic run,
\[
 \Pr_\pi\{\text{ever certifies any clean source}\mid U,Y\}\leq\delta.
\]
The same bound applies to ever actively excluding a clean source under the
confidence gate.  In the exact rotating null, each identity has a correct
3-to-1 majority and \(p_1=\cdots=p_4=1/4\); any certificate is therefore a
false action-target certification and has per-run probability at most \(\delta\).
These conclusions depend on the declared synthetic construction, do not cover
empirical quarantine, and do not become an across-experiment familywise
guarantee when runs are repeated.

\section{Natural Complete-Panel Stress Test}
\label{sec:supp-natural}

\begin{figure*}[t]
\centering
\includegraphics[width=\textwidth]{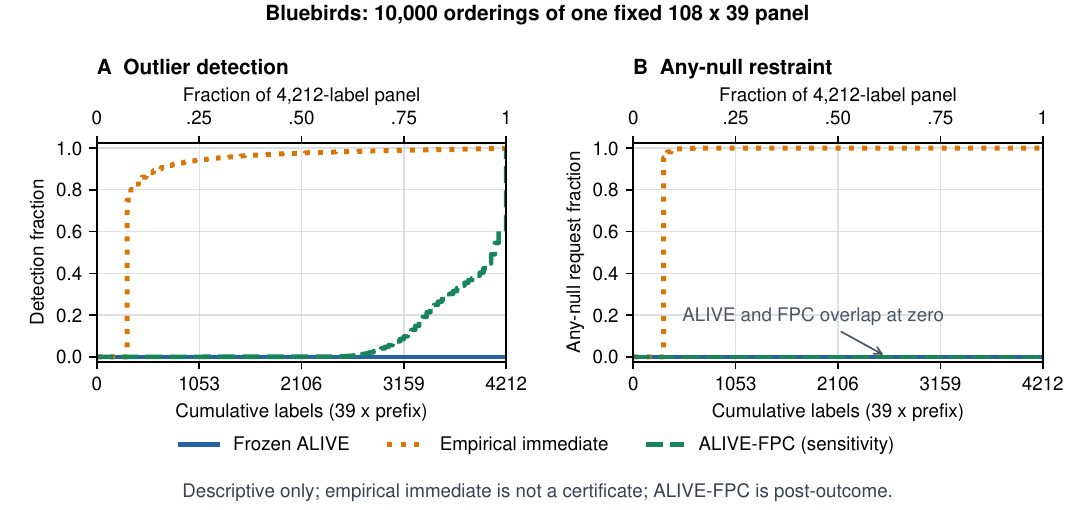}
\caption{Descriptive restraint--power trajectories across 10,000 order
randomizations of one fixed \(108\times39\) Bluebirds panel, not independent
panels.  Panel A is cumulative outlier worker--randomization detection; Panel B
is the fraction of randomizations with any null-worker request.  Empirical
immediate is a descriptive comparator, not a statistical certificate; the
Serfling/census curve is an exploratory sensitivity analysis specified after
the base result, not independent confirmation.}
\label{fig:supp-natural-cost-power}
\end{figure*}

\subsection{Base complete-panel audit}

The public Bluebirds data released with CUBAM \cite{welinder2010crowds} form a
complete binary matrix of 108 tasks by 39 workers.  We use tasks as shared
identities and workers as sources; all 4,212 labels are retained.  Odd panel
size and binary labels give a unique strict majority on every task, hence
\(A=U\) and \(M=108\).  The focal worker remains in that majority.  External
ground truth is unavailable to permutation generation, warning, interval,
certificate, and request predicates; it is used only for the separately
labeled descriptive comparison.

Before computing any worker-level or replay outcome, we fixed the dataset
revision, complete-panel protocol, analysis code, and seed range.  The audit
then consumed 10,000 fixed-seed pseudorandom task permutations from PCG64 seeds
27,010,000--27,019,999.  It retained \(\delta=0.05\), \(\tau=1/S\),
evaluation from prefix eight, two fresh certificate confirmations, and the
remaining-prefix guard.  Label cost at prefix \(n\) is exactly \(39n\).  The
comparison rule instead maps the first empirical warning at or after prefix
eight directly to a persistent-request analogue.  This is intentionally not an ALIVE variant: it tests the consequence of
treating a non-latching warning as a persistent request.  Neither replay includes routing, capacity activation, features, or
training.

The fixed panel has 16 population relative-disagreement outliers and 23 null
workers under Corollary~1.  ALIVE detected 0/160,000
outlier--permutation pairs and produced a null-worker request in 0/10,000 replays.
The empirical analogue detected 160,000/160,000, but issued at least one
null-worker request analogue in 10,000/10,000 replays; across the replays every
worker received at least one analogue request.  Conditional on this fixed panel, the Wilson 95\% upper endpoint for 0/10,000 is
0.000384; it is a Monte Carlo descriptor rather than a replacement for the
theorem's \(\delta\) bound.  These 10,000 permutations probe randomization on one fixed
panel; they are not independent worker panels.

Majority disagreement has Spearman association 0.802 with external worker
error, but majority truth accuracy is only 0.759.  A full-panel binary
Dawid--Skene fit \cite{dawid1979observer} uses deterministic majority
initialization, additive-one smoothing, and a \(10^{-10}\) convergence
criterion; it passes convergence, objective-monotonicity, deterministic-rerun,
and worker-permutation gates in 65 iterations.  Its truth accuracy is 0.889,
and inferred worker quality has Spearman association \(-0.974\) with external
error.  Dawid--Skene is offline, sees every label, and is neither sequential nor
cost matched.  The comparison both supports semantic relevance of disagreement
and demonstrates why no truth interpretation follows from it.

\subsection{Exploratory finite-population completion}

The base radius remains nonzero even at a census.  After observing the base result, we specified one exploratory, non-confirmatory diagnostic,
ALIVE--FPC.  For every ordinary comparable prefix before support exhaustion, it uses
\begin{equation}
 \begin{aligned}
 \rho_{n,|U|}&=1-\frac{n-1}{|U|},\\
 r_n^{\rm FPC}&=\min\!\left\{1,
 \sqrt{\frac{\rho_{n,|U|}\log(2Sn(n+1)/\delta)}{2n}}\right\}.
 \end{aligned}
 \label{eq:supp-fpc-radius}
\end{equation}
All thresholds, seeds, sources, tasks, and ordinary two-fresh and horizon
rules are unchanged.  At support exhaustion, all comparable values have been
seen and the interval is set exactly to \(L_s=U_s=p_s\).  A qualifying exact
census may record one closure request without a duplicate census; it is not an
executed capacity action.

\paragraph{Proposition 2 (FPC coverage and exact closure).}
Suppose the comparable population has unknown size \(M\leq |U|\), and its
order is induced by a uniform permutation of \(U\).  Intervals formed with
Equation~\eqref{eq:supp-fpc-radius} simultaneously cover every source and
ordinary comparable prefix before support exhaustion with probability at least \(1-\delta\).  Hence any
ordinary request for a member of \(\mathcal N\) has probability at most
\(\delta\).  At support exhaustion, an exact-census request for a member of
\(\mathcal N\) is deterministically impossible.

\paragraph{Proof.}
For sampling without replacement, the Serfling inequality gives each tail
probability at most
\(\exp[-2n\epsilon^2/\rho_{n,M}]\)
\cite{serfling1974probability,bardenet2015concentration}.  For fixed \(n\),
\(\rho_{n,M}\) is nondecreasing in \(M\); therefore replacing unknown
\(M\leq|U|\) by \(|U|\) can only widen the interval.  Substitution in the
two-sided bound gives \(\delta/[Sn(n+1)]\).  Summing over sources and prefixes
uses \(\sum_{n\geq1}1/[n(n+1)]=1\), exactly as in Theorem~1.  The
certificate inequalities on simultaneous coverage imply \(s\notin\mathcal
N\); two fresh confirmations and horizon only select a subset.  Once all of
\(U\) is exhausted, \(A\), \(M\), and every \(p_s\) are observed.  The strict
census inequalities then hold exactly if and only if \(s\notin\mathcal N\).
\(\square\)

Before evaluating this diagnostic, implementation checks covered the
conservative-\(M\) direction, closed-form radii, outward rounding, exhaustive
small populations, exact-census predicates, state transitions, horizon, and
deterministic replay.  Because the construction was chosen after observing the
base audit, the result is exploratory rather than independent validation.

ALIVE--FPC closed all 160,000 outlier--permutation pairs with zero null-worker
request in 10,000 replays (Wilson 95\% upper endpoint 0.000384).  Ordinary
two-prefix evidence accounted for 96,117 closures (60.07\%); exact census
accounted for 63,883 (39.93\%).  Median closure prefix was 105/108, and labels
exposed were 4,095/4,212 (97.2\% of the full panel).  Thus FPC reaches finite-population completeness but generally only near
census cost.  Exact
closure is deterministic population identification, not early prediction,
capacity activation, or downstream utility.

\subsection{Exploratory exact-PPR replay}

After the two preceding natural-panel outcomes were known, we specified a
one-shot exact-PPR replay before computing its results.  The analysis is
therefore exploratory and non-confirmatory, although its procedure was fixed
before this replay was evaluated.  The panel, 10,000 PCG64 permutations, seeds
27,010,000--27,019,999, \(N=M=108\), \(S=39\), \(\delta=0.05\),
\(\tau=1/39\), prefix-eight start, two-fresh rule, and two-group horizon are
identical to the parent replay.  Ground truth is not parsed by the order,
interval, certificate, or closure computation.  This instantiates the published
without-replacement PPR confidence-sequence engine
\cite{waudbysmith2020confidence} as a same-target, same-FWER certificate
comparator; it is not a published source-aware or downstream method.

PPR and Serfling are two alternative level-0.05 procedures on the same
source-by-prefix family.  We do not take the union of their action events.  PPR
set membership is checked by exact integer arithmetic with the uniform prior on
total disagreement count.  An ordinary latch requires two consecutive fresh
positive prefixes and is first possible at \(n=9\); the horizon precludes it in
the final two groups.  At \(n=108\), the PPR set is an exact singleton.  A
qualifying event there is recorded only as
\code{exact\_census\_closure}: it is not an ordinary latch, capacity
activation, or early detection.

PPR closed all 160,000 fixed outlier-worker--permutation pairs.  Ordinary
two-fresh evidence accounted for 128,917 closures and exact census for 31,083.
There were zero ordinary null-worker closures and zero census null-worker
closures in 10,000 replays; the ordinary replay-level 0/10,000 Wilson 95\%
upper endpoint is 0.000384.  Against Serfling on exactly paired paths,
PPR was no later on 160,000/160,000, strictly earlier on 128,917 (80.57\%),
equal on 31,083, and later on zero.  The pooled median first-closure prefix was
95 versus 105, corresponding to 3,705 versus 4,095 labels exposed under
complete-panel prefix accounting.  These randomizations probe one fixed small
panel, not independent worker populations; the comparison has no routing,
training, capacity, or downstream utility.  It therefore does not alter the
failed downstream PPR criterion, and the official test set remains unopened.

\begin{table}[t]
\centering
\small
\setlength{\tabcolsep}{2.1pt}
\begin{tabular}{@{}p{0.22\columnwidth}p{0.27\columnwidth}p{0.22\columnwidth}p{0.20\columnwidth}@{}}
\toprule
Rule & Outlier worker$\times$replay closures & Replays with any null closure & Median first closure prefix / labels exposed \\
\midrule
ALIVE (base certificate) &
0/160,000 &
0/10,000 &
--- \\
Empirical immediate &
160,000/160,000 &
10,000/10,000 &
8/312 \\
ALIVE--FPC (exploratory) &
160,000/160,000 &
0/10,000 &
105/4,095 \\
PPR (exploratory) &
160,000/160,000 &
0/10,000 &
95/3,705 \\
\bottomrule
\end{tabular}
\caption{Natural fixed-panel closure diagnostic on the Bluebirds crowd-label
data (108 tasks, 39 workers, and the full 4,212-label panel).  ALIVE--FPC
(exploratory) and PPR were designed after parent outcomes and are
non-confirmatory.  PPR and Serfling are separate level-0.05 procedures; their
actions are not unioned.  Counts describe fixed-panel audit closure behavior.  The median pools ordinary two-fresh requests and exact-census closures; labels exposed are complete-panel prefix accounting (39n), not query savings or downstream utility.}
\label{tab:natural-panel}
\end{table}

\section{Experimental Protocol and Decision Criteria}
\label{sec:supp-protocol}

\subsection{Features, model, and environments}

CIFAR-100 and CIFAR-10 \cite{krizhevsky2009learning} images are encoded once by
an ImageNet-1K-pretrained \cite{deng2009imagenet} ResNet-18
\cite{he2016resnet}.  Each run trains on a seed-specific 10,000-example subset
of a fixed 40,000-example training cache and evaluates on a fixed
10,000-example validation cache.  The head is a one-hidden-layer MLP of width
256, trained with AdamW \cite{loshchilov2019decoupled}, learning rate
\(10^{-3}\), and weight decay \(10^{-4}\).  Candidate count is 512, requested
batch size 256, minimum final batch 32, evaluation interval 25, and proxy
compute anchor 204,000.

\begin{table*}[t]
\centering
\small
\begin{tabular}{@{}p{0.15\textwidth}p{0.17\textwidth}p{0.07\textwidth}
                p{0.25\textwidth}p{0.26\textwidth}@{}}
\toprule
Stage & Environment & Sources & Constructed risk & Audit support\\
\midrule
CIFAR-100 dev & e20-single & 4 & one persistent 0.20 & full 10,000\\
CIFAR-100 dev & e40-single & 4 & one persistent 0.40 & full 10,000\\
CIFAR-100 dev & e60-two-of-five & 5 & two independent persistent 0.60 &
full 10,000\\
CIFAR-100 dev & e80-single & 4 & one persistent 0.80 & full 10,000\\
CIFAR-100 dev & e80-low-overlap & 4 & one persistent 0.80 &
intersection 250; ordinary support full\\
\midrule
CIFAR-10 validation & e20/e40/e80-single & 4 & one persistent
0.20/0.40/0.80 & full 10,000\\
CIFAR-10 validation & e60-two-of-five & 5 & two independent persistent 0.60 &
full 10,000\\
CIFAR-10 null & exact-rate-symmetric-null & 4 &
one rotating wrong source per identity; every \(p_s=0.25\) & full 10,000\\
\bottomrule
\end{tabular}
\caption{Experimental environments.  A persistent rate changes each designated
source label independently at an identity-fixed rate.}
\label{tab:supp-environments}
\end{table*}

In the exact null, source \(s=i\bmod4\) receives \((y_i+1)\bmod K\) and the
other three receive \(y_i\).  Each source disagrees on exactly 2,500
identities; no harmful source is designated.  Null outcomes are never pooled
into ordered-risk utility.

\subsection{Utility estimand and inference}

For metric \(g\), environment \(e\), method \(m\), seed \(z\), and
\(b_1<\cdots<b_4\), normalized AUBC is
\[
 A_{emz}=\frac1{b_4-b_1}
 \sum_{k=1}^{3}\frac{g_{emz}(b_k)+g_{emz}(b_{k+1})}{2}
 (b_{k+1}-b_k).
\]
Budgets are \(0.05,0.10,0.15,0.20\).  The predecessor, ALIVE--CBE,
ALIVE--CBR, Full-Consensus, and switch diagnostics use seeds 40--49; the
separate PPR replication uses previously unused seeds 60--69.  Accuracy is
primary and macro-F1 descriptive.  A contrast is computed within
environment--seed and then averaged equally across the fixed environments.  The
analysis unit is the seed aggregate (\(n=10\)), not a budget, environment,
or stored run.

The reported two-sided sign-flip calculation enumerates all \(2^{10}\)
coordinate-wise signs.  It is exact only under joint sign-exchangeability of
the paired seed effects \cite{ernst2004permutation}; fixed seeds and
deterministic cache splits do not establish that condition.  We therefore
call these conditional enumerated reference values.  Paired-\(t\) and
bootstrap intervals are descriptive.  Failure to reject is never called
equivalence or non-inferiority.

The ALIVE--CBE Holm family contains comparisons against the coupled-action
predecessor, empirical quarantine, empirical agreement, and uniform sampling;
routing-only is outside that family.  ALIVE--CBR has a separate three-comparison
family against standalone CBR, empirical quarantine, and routing-only.  The
later audit-only analysis adds no hypothesis: all contrasts involving it are
outcome-informed, descriptive, and outside Holm.  Full-Consensus and the switch
diagnostics belong to neither family.  The PPR replication likewise has a
separate conjunctive criterion, and its conditional sign-flip value is not
Holm-adjusted.

\paragraph{Exploratory audit-only protocol.}
The diagnostic retained the ALIVE--CBR audit/controller and adaptive 12.5\%/25\%
shadow-warning audit, but forced uniform incumbent allocation, disabled source
action, and made requests and activation impossible.  Its fixed grid crossed
e20/e40/e60/e80, four budgets, and seeds 40--49 (160 validation runs).  The seed
cluster after equal-environment aggregation remained the analysis unit.  It
was specified before its own outcomes but after the ALIVE--CBR results, so it can
only decompose the observed chain descriptively and cannot reopen either the
failed endpoint or the original Holm family.

\paragraph{PPR replication protocol.}
The PPR replication crosses e20/e40/e60/e80, four budgets, seeds 60--69, and
paired ALIVE--PPR/full and routing-only methods (320 utility runs).  The exact
rotating null crosses the same budgets and seeds for the full method only
(40 runs).  Ten equal-environment seed aggregates are the primary units.  Both
methods share the exact PPR prefix and all pre-action behavior; only the full
method may execute certified exclusion.  The reused validation cache and all
constructions have known \(N=M=10{,}000\) and a strict majority at every
identity.

An initial evaluation used the same scientific design, but its analyzer named
the comparator generically while the pre-specified criterion required the
explicit routing-only comparator.  It therefore produced no eligible primary
estimate.  The fresh-seed replication corrected only that interface and was
fixed before its own outcomes.  It is not an external replication, and the
official test set remained unopened.

\subsection{Decision criteria}

\paragraph{ALIVE--CBE development screen.}
The screen required 600 complete, finite, in-budget runs; zero
selected-ineligible or certificate-reopening events; inclusion of all common
true-certifying predecessor and ALIVE--CBE curves; at least 35\% earlier mean
first certification in e40 and e60; a positive mean against the predecessor
with positive leave-one-environment-out means; and a mean against empirical
quarantine above \(-0.0015\).  Missing common timing support counted as failure.

\paragraph{ALIVE--CBE cross-dataset criterion.}
The 960 ordered-risk and 160 null runs had to satisfy integrity and mechanism
checks, e80/e60 detection and e20 abstention, a positive Holm-rejected contrast
against the predecessor with positive leave-one-environment-out effects,
empirical-quarantine price bounds, a positive contrast against uniform sampling
in every environment, and zero certified actions under the exact null while the
empirical comparator produced at least one false isolation.

\paragraph{ALIVE--CBR criterion.}
The 640 ordered-risk and 120 null runs had analogous integrity and mechanism
requirements.  The contrast against standalone CBR had to be positive,
Holm-rejected, positive in e40/e60/e80, and above \(-0.0015\) in e20.
Routing-only had to be positive, empirical quarantine had to satisfy the same
price bounds, and the null had to show zero certified ALIVE--CBR actions but an
empirical false isolation.

\paragraph{Full-Consensus validity.}
Full-Consensus had no performance threshold.  Reporting required 160 unique,
complete, finite, in-budget runs; exact input identities; strict reconstruction
of alignment, majority targets, selected indices, and ledger counts; and a
complete matched ALIVE--CBE comparator.  Every valid result direction was then
reported.

\paragraph{Adaptive-Switch diagnostic criteria.}
The exploratory criteria jointly required complete integrity, e20 parity with
ALIVE--CBR, zero e20 actions, the intended one-decision delayed switch in
higher-risk environments, exact-null abstention, an ALIVE--CBE improvement of at
least \(0.001\) with conditional sign-flip value below \(0.05\), a positive
high-risk contrast against ALIVE--CBR, and a positive all-risk contrast against
CBR.  Failure of any clause precluded further official evaluation.

\paragraph{ALIVE--PPR replication criterion.}
The conjunctive criterion comprised: (i) exactly 320 utility and 40 null runs
that were complete, finite, issue-free, and in budget; (ii) known
\(N=10{,}000\), complete support, strict majorities, a uniform
without-replacement prefix, and exact-integer PPR reconstruction; (iii) matched
pre-action paths, zero e20/null action, and no clean-source certificate;
(iv) exact harmful-set exclusion on all high-risk curves, PPR no later than the
Hoeffding engine on at least 90\%, strictly earlier on at least 50\%, and no
worse median evidence count in each high-risk environment; and (v) full minus
routing-only accuracy AUBC at least \(+0.001\), conditional two-sided sign-flip
\(p<0.05\), a strictly positive mean in each of e40/e60/e80, a nonnegative
pooled high-risk 5\%-budget mean, positive leave-one-high-risk-environment-out
means, and exact e20 parity.  All clauses had to hold jointly.  Thresholds could
not be revised after observing outcomes, and the official test set remained
closed on failure.

\section{Complete Experimental Results}
\label{sec:supp-outcomes}

\subsection{Coupled-action predecessor and selector study}

The coupled-action predecessor has 1,400/1,400 valid, finite,
budget-compliant runs and zero active
quarantine training violations.  Severe one- and two-source failures were
certified/quarantined in 10/10 seed aggregates; e20 abstained in 10/10.  Because
e20 fixes \(p_s=0.20<\tau=1/4\), that abstention is the declared action-target
boundary, not a failure to detect an alternative inside the target.
Nevertheless, the predecessor minus empirical agreement accuracy AUBC was
\(-0.0048277\), descriptive paired-\(t\) 95\% interval
\([-0.0056235,-0.0040318]\), with all ten seed effects negative.  The
conditional enumerated value was \(0.001953\), Holm-adjusted to \(0.0078125\),
but in the unfavorable direction.  The predecessor exceeded routing-only by
\(0.0072773\).  The pre-specified headline criterion was therefore not met: the mechanism
and integrity checks passed, whereas the required positive primary effect and
every positive leave-one-environment-out effect failed.

The selector-factor study has 1,000/1,000 valid, finite, in-budget runs.  Mean
accuracy AUBC
was \(0.499892\) for CBR, \(0.470603\) for complete historical ALIVE,
\(0.468894\) for CBE, \(0.460334\) for random, and \(0.411315\) for entropy.
Complete ALIVE minus CBR was \(-0.029289\), negative for all ten seeds and
every environment--budget cell; complete ALIVE minus CBE was only
\(+0.001709\).  This result motivated the ALIVE--CBR variant but is not evidence for its
effectiveness.

\subsection{ALIVE--CBE development screen}

The strict matrix has 600/600 complete finite runs, zero budget overruns,
selected-ineligible violations, or certificate reopenings.  On the 40 common
true-certifying curves per environment, mean first certification moved from
decision 21.9 to 12.3 in e40 (43.84\% earlier) and from 12.8 to 7.3 in e60
(42.97\% earlier); all 40/40 curves remained true-certifying in both.

ALIVE--CBE minus the predecessor mean accuracy AUBC was \(+0.0023127\), range
\([-0.0001200,+0.0065800]\) over ten seed aggregates.  All
leave-one-environment-out means were positive:
\(+0.0031871,+0.0021100,\allowbreak +0.0023175,\allowbreak
+0.0022954,\allowbreak +0.0016533\) when excluding
e20, e40, e60, e80, and e80-low-overlap, respectively.  ALIVE--CBE minus empirical
quarantine was \(-0.0007240\), above the pre-specified \(-0.0015\) tolerance.  Every development condition passed, permitting the pre-specified
cross-dataset validation stage.

\subsection{ALIVE--CBE cross-dataset validation}

The ordered-risk package has 960/960 and the exact null 160/160 complete
finite in-budget runs, with zero selected-ineligible or reopening events.
ALIVE--CBE minus the predecessor accuracy AUBC was \(+0.0047042\), range
\([+0.0024958,+0.0069042]\), descriptive paired-\(t\) 95\% interval
\([+0.0035212,+0.0058871]\), and positive for all ten seed aggregates.
The conditional sign-flip value was \(0.001953\), Holm-adjusted to
\(0.0078125\).  All leave-one-environment-out means were positive:
\(+0.0063961,+0.0047567,+0.0037422,+0.0039217\).

\begin{table}[t]
\centering
\small
\begin{tabular}{@{}lr@{}}
\toprule
ALIVE--CBE contrast (accuracy AUBC) & Mean\\
\midrule
minus predecessor & \(+0.0047042\)\\
minus empirical quarantine & \(-0.0002371\)\\
minus empirical agreement & \(+0.0001558\)\\
minus uniform & \(+0.0078183\)\\
minus routing-only & \(+0.0051717\)\\
\bottomrule
\end{tabular}
\caption{ALIVE--CBE equal-environment seed-clustered means.}
\label{tab:supp-cbe-effects}
\end{table}

The empirical-quarantine contrast had descriptive paired-\(t\) interval
\([-0.0004144,-0.0000598]\), satisfying the price guard; routing-only had
interval \([+0.0033102,+0.0070332]\).  The e80, e60, and e20 joint mechanisms
qualified in 10/10 seeds each.  Under the exact null, all 40 ALIVE--CBE curves
had zero certificates, exclusion requests, and active quarantines, whereas empirical
quarantine falsely isolated on all 40 curves.  ALIVE--CBE minus empirical quarantine there was \(-0.0288483\).

The conditional headline nevertheless failed.  Nine of ten conditions passed,
but ALIVE--CBE minus uniform was \(-0.0003717\) in e20, violating the
requirement of
a positive effect in \emph{every} environment (e40 \(+0.0071233\), e60
\(+0.0136117\), e80 \(+0.0109100\)).  Accordingly, the cross-dataset headline criterion was not met; the
strong
primary comparison does not erase that pre-specified failure.

\subsection{ALIVE--CBR: component improvement without a system-level win}

ALIVE--CBR has 640/640 ordered-risk and 120/120 null runs, all complete,
finite, and
in budget, with zero violations or reopenings.  All e80/e60/e20 mechanisms
qualified in 10/10 seeds.  ALIVE--CBR minus CBR accuracy AUBC was
\(+0.0019538\), range \([-0.0025542,+0.0063792]\), descriptive paired-\(t\)
95\% interval \([+0.0000330,+0.0038745]\), with seven wins and three losses.
The conditional value \(0.048828\) became \(0.097656\) after its separate Holm
adjustment.  Environment means were e20 \(-0.0014233\), e40
\(-0.0024200\), e60 \(+0.0062083\), and e80 \(+0.0054500\).

ALIVE--CBR minus routing-only was \(+0.0019346\), interval
\([+0.0011720,+0.0026971]\), positive in all ten seeds and Holm-adjusted
\(p=0.005859\).  ALIVE--CBR minus empirical quarantine was \(-0.0000058\),
interval
\([-0.0002009,+0.0001892]\); this is not evidence of equivalence.  The null
had zero ALIVE--CBR certificates, requests, activations, violations, or
reopenings, while the empirical ablation falsely isolated.

Eight of nine ALIVE--CBR headline conditions passed.  The combined primary
condition
failed because the Holm-adjusted value exceeded \(0.05\) and e40 was negative.
The pre-specified system-level criterion was therefore not met; no headline
claim or threshold revision was made.

\begin{figure*}[t]
\centering
\includegraphics[width=\textwidth]{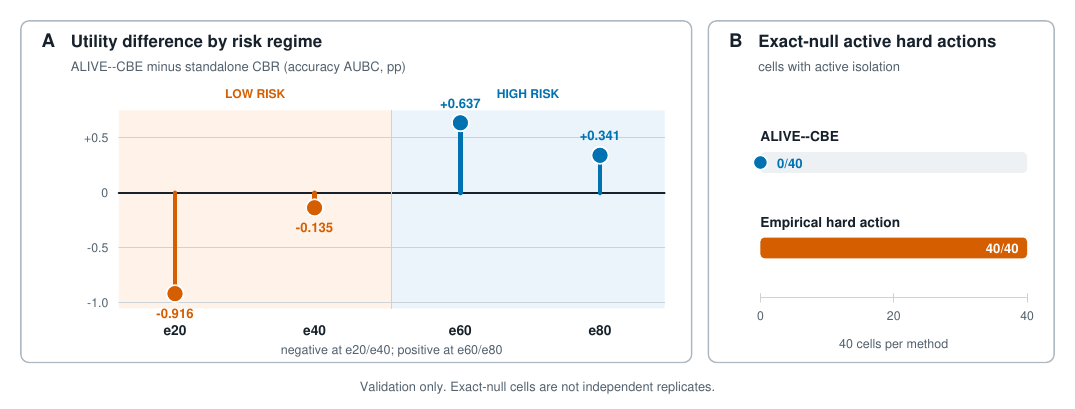}
\caption{Matched restraint--utility evidence.  Left: the ALIVE--CBE versus
CBR contrast changes sign with risk.  Right: exact-null active actions; cells
are repeated diagnostics, not independent replicates.}
\label{fig:safety-utility}
\end{figure*}

\subsection{Exploratory audit-only decomposition}

All 160/160 runs completed in budget.  The behavior audit reconstructed 15,400
decisions and 1,021,396 audit slots, including 796 decisions at the 25\% shadow
audit rate.  It found zero non-incumbent routes, requests, activations,
allocation violations, cheap-evaluation events, and paired-audit mismatches.
Thus the control retained the audit opportunity cost while removing both source
actions.

Accuracy-AUBC seed-cluster means were audit-only minus standalone CBR
\(-0.0022742\) (\(-0.2274\) pp), routing-only minus audit-only \(+0.0022933\)
(\(+0.2293\) pp), and ALIVE--CBR minus audit-only \(+0.0042279\)
(\(+0.4228\) pp).  Together with the original ALIVE--CBR minus
routing-only effect \(+0.0019346\) (\(+0.1935\) pp), the chain closes to the
unchanged endpoint ALIVE--CBR minus CBR \(+0.0019538\) (\(+0.1954\) pp).
The endpoint still failed its original Holm test (adjusted \(p=0.097656\)); no
new contrast belongs to a Holm family.

\subsection{Full-Consensus: valid boundary with lower utility}

The validity checks passed on 160/160 finite in-budget runs.  Exact aligned
feature reconstruction found 6,048,200 proposals, 5,896,768 unique identities,
151,432 replacement collisions (2.50375\%), 25,657,120 source-label queries,
and strict-majority availability for every proposal: zero abstentions and zero
simulator-truth disagreements.  Training used 3,007,810 selected examples over
11,830 optimizer steps.

Full-Consensus mean accuracy AUBC was \(0.8177667\), versus \(0.8237479\)
for ALIVE--CBE.
Full-Consensus minus ALIVE--CBE was \(-0.0059813\), with descriptive
paired-\(t\) interval
\([-0.0081912,-0.0037714]\), bootstrap interval
\([-0.0079196,-0.0042850]\), zero wins in ten, and conditional descriptive
sign-flip value \(0.001953\).  Environment effects were
\(-0.0011550,-0.0066817,-0.0099983,-0.0060900\) for e20/e40/e60/e80.
Macro-F1 differed by \(-0.0072104\), interval
\([-0.0098088,-0.0046120]\), again zero wins.  These are mandatory
post-diagnostic results, not a failed promotion test.

Ledger totals across all Full-Consensus runs were 513,142.4 source,
302,410.0 cheap
evaluation, 256,571.2 maintenance, and 3,007,810.0 training units, totaling
4,079,933.6; fine evaluation and refresh were zero.  Full consensus is thus
neither a free oracle nor a superior boundary in this matrix.

\subsection{Adaptive-Switch diagnostic}

All 160 adaptive and 40 null runs passed the behavior audit, remained in budget,
and had no selected-ineligible events.  In high-risk environments, ranking was
CBR at the first certificate decision, entropy began exactly one decision
later, all charges were reconstructed, and the switch was monotone.  The exact
null produced no certificates, exclusions, or entropy switches.

The e20 parity check initially reported 40 selection and 40 metric mismatches.
A fieldwise audit showed that selected indices, source IDs, trace lengths,
ledger events, controller states, and structural diagnostics were identical in
all matched cells.  Differences were confined to model-dependent floating-point
diagnostics: Adaptive-Switch minus ALIVE--CBR e20 accuracy had mean
\(-2.496\times10^{-6}\) and range \([-0.0018000,+0.0013000]\); macro-F1 mean
was \(-4.773\times10^{-6}\), and compute and overhead metrics were exact.  The
most plausible explanation is GPU numerical nondeterminism because deterministic
PyTorch/cuDNN/CUBLAS execution was not enforced.  Thus the original parity
label was overinclusive; it is not evidence that different examples or sources
were selected.

This clarification does not change the scientific decision.  Three independent
criteria failed: the pre-specified exact-parity clause; the primary conditional
sign-flip requirement (mean versus ALIVE--CBE \(+0.0011783\), but
\(p=0.326172\)); and the high-risk contrast against ALIVE--CBR
(\(-0.0012650\)).  The all-risk contrast against CBR was positive
(\(+0.0009975\)), while the integrity, delayed-switch, e20-abstention, and null
criteria passed.  Adaptive-Switch therefore remains an exploratory diagnostic,
and the official test set was not evaluated.

\subsection{Fixed-Switch diagnostic and cost disclosure}

Fixed-Switch's 160/160 validation runs passed the no-metric behavior audit
with zero
issues or budget violations.  The first entropy decision was exactly 8 in
every run; decision counts at the four budgets were 36, 71, 106, and 141.
The auditor reconstructed exactly 12,880 candidate-forward events, 9,708
quarantine-request steps, and 9,708 active-quarantine steps, with no
eligibility violation.

Fixed-Switch accuracy AUBC was \(0.8226146\) and macro-F1 AUBC \(0.8208347\).
Adaptive-Switch minus Fixed-Switch accuracy AUBC was \(+0.0023117\), conditional
descriptive sign-flip value \(0.0390625\), with environment effects e20
\(+0.0088650\), e40 \(+0.0000567\), e60 \(+0.0017417\), and e80
\(-0.0014167\).  This has no confirmatory status or multiplicity adjustment.
Fixed-Switch minus ALIVE--CBE was \(-0.0011333\) (\(p=0.185547\));
Fixed-Switch minus ALIVE--CBR was
\(-0.0032679\) (\(p=0.001953\), all ten negative); and Fixed-Switch minus CBR was
\(-0.0013142\) (\(p=0.201172\)).  The last contrast varied sharply:
e20 \(-0.0103183\), e40 \(-0.0031517\), e60 \(+0.0053333\), and e80
\(+0.0028800\).

Across 160 runs, mean cheap-evaluation, training, overhead, and total proxy
costs were 2,060.8, 22,527.75, 2,971.68625, and 25,499.43625.  Mean overhead
ratio was 0.1145392 and mean compute-to-budget ratio 0.9999688.  These costs and all comparisons are descriptive; they do not alter the
Adaptive-Switch conclusions or justify official evaluation.

\subsection{ALIVE--PPR: lower evidence cost without satisfying the headline criterion}

The initial PPR evaluation completed 320 utility and 40 null runs, but its
analysis interface did not name the routing-only comparator in the exact form
required by the pre-specified criterion.  It therefore yielded no eligible
primary estimate and is not used for scientific interpretation.  The
fresh-seed replication retained the method, PPR calculation, controller,
trainer, cost model, comparator, environments, budgets, thresholds, and all
numeric criteria; only the analyzer--comparator interface, provenance fields,
and seed domain changed.

The replication completed all 320 utility and 40 null runs; every trajectory
was finite, in budget, and valid for the known-population PPR construction.
All 160 matched pairs had identical pre-action selections and ledgers.  All e20
full curves and all null curves had zero certificates, exclusion requests, and
active exclusions, with no clean-source violation.  Every one of the 120
high-risk full curves certified exactly the constructed harmful set and
activated exclusion.  PPR separation was no later than Hoeffding on 120/120 and
strictly earlier on 100/120 curves.  Median PPR versus Hoeffding evidence counts
were 96 versus 304 in e40, 62 versus 171 in e60, and 48 versus 48 in e80.

\begin{table}[t]
\centering
\small
\setlength{\tabcolsep}{2.5pt}
\begin{tabular}{@{}p{0.24\columnwidth}p{0.12\columnwidth}p{0.53\columnwidth}@{}}
\toprule
Criterion & Result & Evidence \\
\midrule
Integrity & pass & 320 utility + 40 null runs complete and valid \\
PPR validity & pass & 360/360 known-population valid; zero ties \\
Path matching & pass & 160/160 matched; e20/null zero action \\
Mechanism & pass & 120/120 correct; no later 120/120, earlier 100/120 \\
Utility & fail & mean \(=+0.224\) pp, \(p=0.001953\); e40 \(=-0.056\) pp \\
Joint & fail & e40 clause fails; official test remains unopened \\
\bottomrule
\end{tabular}
\caption{Pre-specified ALIVE--PPR replication criteria.  The utility clause fails despite a positive pooled mean because e40 is negative; the joint criterion therefore fails.}
\label{tab:ppr-gate-summary}
\end{table}

\begin{table}[t]
\centering
\small
\setlength{\tabcolsep}{2.4pt}
\begin{tabular}{@{}lrrrrr@{}}
\toprule
Env. & 5\% & 10\% & 15\% & 20\% & AUBC \\
\midrule
e20 & \(+0.000\) & \(+0.000\) & \(+0.000\) & \(+0.000\) & \(+0.000\) \\
e40 & \(+0.012\) & \(+0.008\) & \(-0.219\) & \(+0.077\) & \(-0.056\) \\
e60 & \(+0.736\) & \(+0.638\) & \(+0.625\) & \(+0.428\) & \(+0.615\) \\
e80 & \(+0.132\) & \(+0.380\) & \(+0.337\) & \(+0.442\) & \(+0.335\) \\
\bottomrule
\end{tabular}
\caption{ALIVE--PPR replication mean paired accuracy differences (full minus routing-only, percentage points; seeds 60--69).  AUBC is the normalized trapezoidal integral; positive values favor the full method.}
\label{tab:ppr-env-budget}
\end{table}

The primary accuracy-AUBC difference, full minus routing-only, was
\(+0.223541\) pp: all ten seed aggregates were positive, the two-sided
conditional exact sign-flip value was \(0.001953125\), and the descriptive
paired-\(t\) 95\% interval was \([+0.1468,+0.3003]\) pp.  Environment means
were exactly \(0\) in e20, \(-0.055500\) pp in e40 (3 wins, 7 losses),
\(+0.615000\) pp in e60, and \(+0.334667\) pp in e80.  The pooled high-risk
5\%-budget mean was \(+0.293334\) pp and every leave-one-high-risk-environment
mean was positive.  However, the pre-specified utility criterion required a
strictly positive mean in every high-risk environment; the negative e40 result
violated that clause.  The conjunctive headline criterion was therefore not
met.

Macro-F1 was descriptive rather than part of the pre-specified criterion.  Its
aggregate difference was \(+0.227030\) pp, with e20 \(0\), e40
\(-0.043399\) pp, e60 \(+0.605671\) pp, and e80 \(+0.345848\) pp.  These
secondary results do not alter the failed e40 clause.  No threshold revision,
official-test access, external-replication claim, or deployment claim is made;
the PPR study remains a validation-adaptive, fresh-seed contract-repair
replication.

\section{Unavailable Resource Comparisons}
\label{sec:supp-resource}

Alternative-resource integration requires a positive-width coordinate support
common to every method--seed curve in an environment.  The selector-factor study failed this
condition for its global resource comparison.  The ALIVE--CBE development
analysis included all 600 runs but found zero common-width trainer-core-time
support in e60.  The ALIVE--CBR analysis found the same limitation in all four
environments.  Consequently, no aggregate estimate is reported for those axes; partial
environment summaries or support redefinitions would change the estimand.

The declared-budget results and exact ledgers remain available.  Synchronized
trainer-core time, where defined, covers only trainer entry through selection,
training, and final evaluation with end synchronization; it excludes setup,
launch, and serialization and is not complete-job wall time.  The 16 fixed-trajectory
repricings of fine-evaluation and maintenance costs leave trajectories fixed
and are descriptive feasibility views, not evidence of behavior under an
actual alternate-price rerun.

\section{Study Sequence and Outcome Isolation}
\label{sec:supp-study-sequence}

The experiments were developed sequentially on a reused validation cache, so
later variants are validation-adaptive rather than members of one independent
confirmatory family.  Table~\ref{tab:supp-study-sequence} reports the sequence
with reader-facing names.  Internal package identifiers are included only to
connect the manuscript to the anonymous code package; they are not method
versions and are not used elsewhere in the exposition.

\begin{table*}[t]
\centering
\small
\setlength{\tabcolsep}{3.5pt}
\begin{tabular}{@{}p{0.21\textwidth}p{0.10\textwidth}p{0.31\textwidth}p{0.29\textwidth}@{}}
\toprule
Study & Package ID & Role & Outcome status\\
\midrule
Coupled-action predecessor and selector study & \code{R15} & Initial controller
and incumbent-selection evidence & Mechanism checks passed; utility criterion
failed; selector favored CBR\\
ALIVE--CBE & \code{R16} & Development screen followed by cross-dataset
validation & Development screen passed; cross-dataset criterion failed in e20
versus uniform\\
ALIVE--CBR & \code{R17} & Matched CBR-incumbent controller & Persistent-action
increment was positive; adjusted system-level criterion versus CBR was not met\\
Full-Consensus & \code{R18} & Post-diagnostic aggregation boundary & Valid
execution; lower utility than ALIVE--CBE\\
Adaptive-Switch & \code{R19} & Outcome-informed ranking diagnostic & Valid
execution; parity, primary significance, and high-risk improvement criteria
failed\\
Fixed-Switch & \code{D8} & Pre-fixed schedule diagnostic & Descriptive only;
no promotion criterion\\
Initial PPR evaluation & \code{R20} & Same scientific design as the PPR
replication & Analysis-interface mismatch produced no eligible primary estimate\\
PPR replication & \code{R20.1} & Fresh-seed repair of that interface &
Mechanism criteria passed; the e40 utility clause failed\\
\bottomrule
\end{tabular}
\caption{Study sequence, package mapping, and outcome status.  All unfavorable
results are retained.  The official test set remained unopened.}
\label{tab:supp-study-sequence}
\end{table*}

ALIVE--CBE was fixed before its development outcomes.  ALIVE--CBR was specified
from the selector study and fixed before its own model outcomes, with a separate
Holm family.  Full-Consensus and the audit-only control were introduced after
related results were available and are explicitly descriptive.  Adaptive-Switch
was motivated by opened validation evidence but fixed before its own outcomes;
Fixed-Switch was specified independently of those outcomes.  The initial PPR
evaluation was fixed before execution.  The PPR replication was specified only
after its analyzer--comparator interface defect was observed; it retained the
scientific design and numeric criteria while using fresh run and audit seeds.
These controls provide local outcome isolation, not an externally timestamped
preregistration or an independent population.

\section{Reproducibility and Provenance}
\label{sec:supp-reproduction}

The anonymous code package contains a SHA-256 manifest that binds the protocol,
configuration, code, run grid, behavior-audit outputs, statistical analyses,
and generated tables/figures for each package ID in
Table~\ref{tab:supp-study-sequence}.  The manuscript omits the raw digest list:
the machine-readable manifest is the authoritative byte-level index and can be
verified directly from the extracted package.  No absolute paths, host names,
home directories, or machine timestamps are required.

Reproduction follows the same five-stage workflow for each study:
(i) verify the packaged manifest and configuration; (ii) execute the complete
method--environment--budget--seed grid; (iii) reconstruct trajectories and
ledger events without reading outcome metrics; (iv) compute the pre-specified
statistical summaries; and (v) evaluate the corresponding decision criteria.
Unknown IDs, duplicate or incomplete grid cells, non-finite metrics, type
mismatches, budget violations, or digest mismatches invalidate the affected
analysis.  Resume operates at whole-run granularity: an incomplete run restarts
from its seed and replaces the partial trace.

Code-only validation covers compilation and linting, unit tests for selection,
allocation, ledgers, replay, and the exact rotating-null construction.  These
tests verify implementation invariants; they are not conditioned on obtaining
a favorable scientific result.  For the PPR replication, provenance also binds
the implementation diff from the initial evaluation, the fresh seed domains,
and the explicit routing-only comparator used by the unchanged scientific
criterion.

\section{Limitations}
\label{sec:supp-limitations}

The theorem assumes fixed common support and potential labels, with probability taken over one ideal audit permutation. Its simultaneous guarantee applies within a run and does not extend familywise across repeated seeds, datasets, or deployments. The estimand excludes source-only records, temporal drift, call-dependent noise, and identities without a strict majority. Since \(U=\bigcap_s U_s\), adding sources may reduce common support; partial overlap or changing source membership requires a revised estimand and certificate.

\textsc{Certified} means satisfying the pre-specified absolute disagreement and relative peer-separation criteria, not that a source is necessarily incorrect or harmful. The threshold \(\tau\) must be chosen before label inspection, and (1/S) is an experimental setting rather than a universal corruption rate. Consensus-based certification may favor a shared incorrect majority or flag a correct minority, while monitoring and fallback mechanisms provide operational safeguards rather than downstream loss guarantees.

The experiments provide controlled evidence using synthetic persistent perturbations, fixed CIFAR features, a small MLP head, four proxy budgets, and ten paired seeds per study. CIFAR-10 serves as cross-dataset validation rather than a natural multi-provider deployment. The study sequence is validation-adaptive, the official test cache remains unopened, and the proxy ledger is specific to the evaluated trajectories and costs. The Full-Consensus, Adaptive-Switch, PPR, audit-only, and Bluebirds analyses should therefore be interpreted as complementary diagnostic or exploratory evidence. The PPR replication improves the analysis interface but is not external confirmation and does not satisfy the pre-specified e40 utility condition; Bluebirds provides only a small complete panel without routing or training, where the base certificate has no detection power and the direct certificate remains unevaluated.

\bibliography{references}